\documentclass[letterpaper,12pt]{article}
\pdfoutput=1
\usepackage[all]{xy}
\usepackage{amsmath}
\usepackage{amssymb}
\usepackage{amsthm}
\usepackage{amsopn}
\usepackage{mathrsfs}
\usepackage[pdftex]{color,graphicx}
\newtheorem{thm}{Theorem}[section]

\newtheorem{conjecture}[thm]{Conjecture}

\DeclareMathOperator{\diag}{diag}
\DeclareMathOperator{\num}{num}
\DeclareMathOperator{\den}{den}

\numberwithin{equation}{section}
\author{Steven Wegmann \\ Nuance Communications \\ Mobile Research}
\title{Approximations to the MMI criterion and their effect on lattice-based MMI.}
\date{}
\begin{document}
\maketitle 
\section{Introduction}

Although maximum mutual information (MMI) training has been used for hidden Markov model (HMM) parameter estimation for more than twenty years (\cite{bahl}, \cite{merialdo}, \cite{gopal}, \cite{normandin}, and \cite{valtchev2}), it has recently become an essential part of the acoustic modeling repertoire thanks to the refinements introduced by Woodland and Povey (\cite{woodland2} and \cite{povey2}).  The earliest incarnations of MMI worked well on small vocabulary tasks with small models, for example digit recognition.  However, one can expect to gain 10-20\% in recognition accuracy over standard maximum likelihood methods regardless of the size of the task or the models when using the current methodology, lattice-based MMI.

The machinery of lattice-based MMI consists of a model selection criterion called the MMI criterion and an iterative estimation algorithm called the extended Baum-Welch algorithm.  This machinery is analogous to -- it is in fact based on -- the standard machinery used for maximum likelihood estimation with HMMs, where the model selection criterion is the log-likelihood of the training data and the iterative estimation algorithm is the Baum-Welch algorithm (\cite{baum}).  In both cases the estimation algorithm operates on the space of all possible model parameters by producing a new estimate of model parameters from an original estimate.  Also, both of these estimation algorithms have been designed so that the model selection criterion is larger on the new estimate than it was on the original estimate.  Finally, in both cases the machinery is operated in the same manner: starting from a choice of initial model parameters, we repeatedly apply the estimation algorithm, first to the initial choice, next to the result of this, etc., thereby creating a sequence of model parameters.  In the case of maximum likelihood estimation with the Baum-Welch algorithm the properties of the resulting sequence of model parameters are understood and generally good, while in the case of lattice-based MMI with extended Baum-Welch these properties have never been studied.\footnote{However, the tacit assumption in the literature is that the sequence of model parameters produced by extended Baum-Welch does converge.  For example the notion \emph{weak sense auxiliary functions} in \cite{povey2} appears to depend upon this convergence.} 

\begin{figure}
\centering
\includegraphics[width=1.0\textwidth]{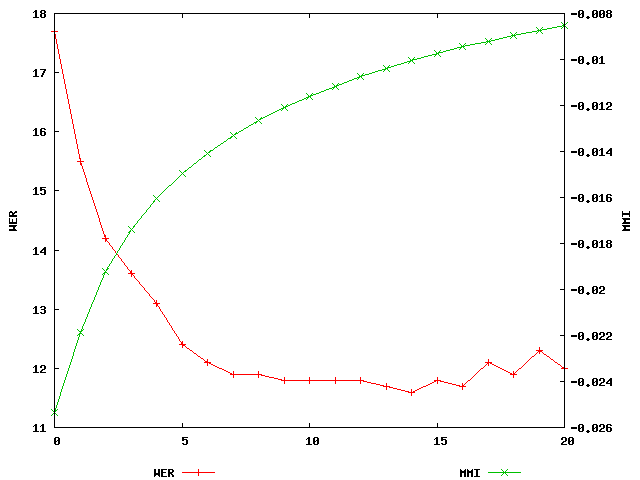}
\caption{The WER on an independent test set and the MMI criterion on the training data during twenty iterations of extended Baum-Welch.  The x-axis gives the extended Baum-Welch iteration, with x=0 being the mle.\label{usual_mmi_graph}} 
\end{figure}

Figure \ref{usual_mmi_graph} is representative of plots that appear in nearly all of the literature on MMI.  We note that the MMI criterion steadily increases over the twenty iterations, while the word error rate (WER) initially decreases, levels out, and then begins a slight upward trend with notable oscillation.  The conventional wisdom has been that this is due to `over training', i.e., that MMI, whether lattice-based or not, somehow over specializes the models to the training data at the expense of recognition performance on more general test data.  Even if one believes this explanation it is worth understanding what the mechanism is that is to blame for this over specialization. Is it a property of the algorithm, extended Baum-Welch, or a property of the model selection criterion, the MMI criterion, or something else entirely that is at the root of the problem?  This the central question that we will address in this paper.

\begin{figure}
\centering
\includegraphics[width=1.0\textwidth]{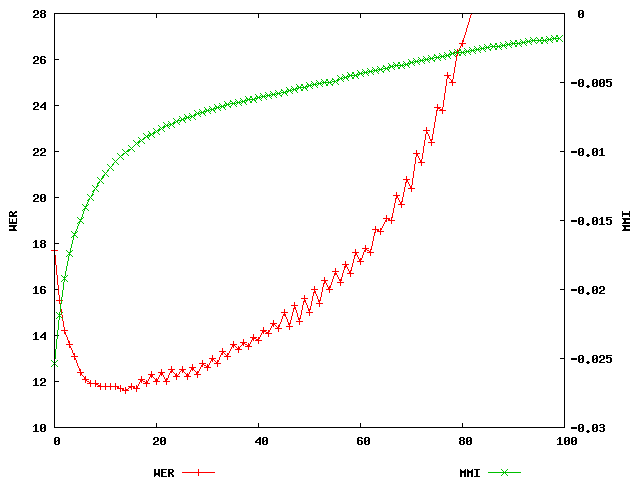}
\caption{The WER on an independent test set and the MMI criterion on the training data during 100 iterations of extended Baum-Welch.  The x-axis gives the extended Baum-Welch iteration, with x=0 being the mle.\label {longer_mmi_graph}} 
\end{figure}

One of the starting points of this research was to investigate what happens if we run many more iterations of extended Baum-Welch than is typical.  The motivation was to investigate whether or not the model parameters are actually converging.  Figure \ref{longer_mmi_graph} extends the results in Figure \ref{usual_mmi_graph} by running eighty more iterations of extended Baum-Welch.  The MMI criterion is by design supposed to be more predictive of recognition performance than the maximum likelihood criterion.  Yet we see that the MMI criterion steadily increases while the corresponding recognition performance falls apart.  Since the MMI criterion has not converged, we also conclude that the models parameters have not converged even after 100 iterations. 

\begin{figure}
\centering
\includegraphics[width=1.0\textwidth]{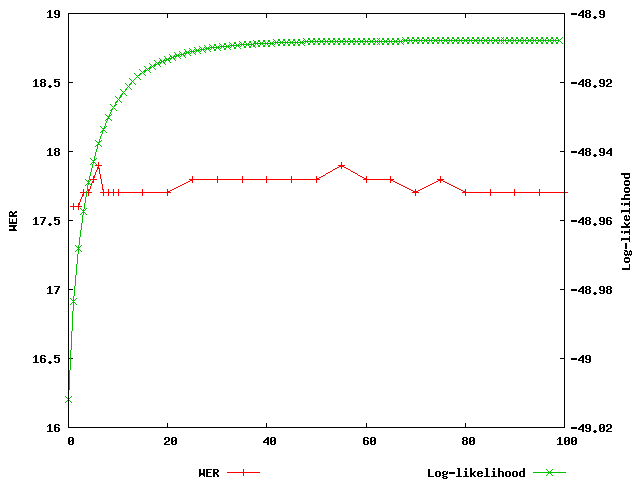}
\caption{The WER on an independent test set and the log-likelihood on the training data during 100 iterations of Baum-Welch.  The x-axis gives the Baum-Welch iteration, with x=0 being initial models.\label {usual_bw_graph}} 
\end{figure}

This is in stark contrast to what happens with maximum likelihood estimation with the Baum-Welch algorithm.  Figure \ref{usual_bw_graph} shows the analogous experiment.  Note that the behavior is much more benign.  The log-likelihood steadily increases, as theory predicts, but it appears to be converging to around \(-48.9\).  Also, while the WER oscillates, the amplitude is very small, and it too appears to be converging to \(17.7\%\).  Note that this, as a practical matter, is much more desirable behavior than what we observed in Figure \ref{longer_mmi_graph}.  There is little to gain by worrying about how long we run maximum likelihood estimation, while one has to be very careful to run lattice-based MMI just the right number of times.  To make matters worse, since the MMI criterion is puzzlingly disconnected from test set recognition performance, we are forced to use recognition performance on a independent validation set to determine when to stop extended Baum-Welch.  Aside from being puzzling, this disconnect between the MMI criterion and general recognition performance leaves one vulnerable to questions about how to construct an adequate validation test set for model selection.  These considerations also make lattice-based MMI more difficult to fit into a fully automatic assembly-line acoustic model factory than maximum likelihood estimation.

So, again, why does lattice-based MMI behave so differently from maximum likelihood estimation?  As we will remind the reader in Section~\ref{notation}, for practical reasons we actually use an approximate version of the MMI criterion.  In lattice-based MMI there are two aspects to this approximation and they are both encapsulated in the use of phone-marked word lattices.  The first approximation occurs in the calculation of the MMI criterion via Bayes' Rule.  Instead of summing over all possible transcriptions of a given training utterance -- which is impossible -- we restrict this sum to the transcriptions that occur in the corresponding lattice.  The second approximation occurs in the calculation of the HMM-based acoustic scores which are inputs for the calculation of the MMI criterion.  Instead of summing over all possible state sequences compatible with a given transcription, as the HMM demands, we restrict this sum to a subset of state sequences that are compatible with the phone-level time boundaries -- the phone-marks -- that occur in the corresponding lattice.  The resulting approximation to what we might call the true MMI criterion is what extended Baum-Welch actually uses as a model selection criterion in lattice-based MMI.

In this paper we will demonstrate that the properties of lattice-based MMI depend on the properties of this approximation.  The usual practice is to first generate the phone-marked word lattices using the mle seed models, then use these fixed lattices throughout multiple iterations of extended Baum-Welch.  This results in the behavior displayed in Figure \ref{longer_mmi_graph}.  In Section~\ref{fixedLats} we will demonstrate that difference between the approximate and true MMI criteria is very small at the mle but steadily increases as the model parameters move away from the mle.  We will also demonstrate that the approximate MMI criterion appears to attain its maximum value not within the model parameter space but, instead, at a point at infinity.  This suggests that model estimation using the approximate MMI criterion is an ill-posed problem.  It also means that the approximate MMI criterion is a terrible choice to perform dual r\^oles, first  as an approximation to the MMI criterion and second as a model selection criterion, since by design it will produce estimates for the model parameters that are far from the mle where it is no longer related to the MMI criterion.   Extended Baum-Welch obliges these properties by producing a sequence of parameters that heads to a point at infinity with steadily increasing approximate MMI criterion.  In this respect, the algorithm, extended Baum-Welch is blameless, it is merely obeying the pathological demands of the approximate MMI criterion.

In Section~\ref{regenWordPhoneLats} we explore what happens if we use a much better approximate MMI criterion.  We accomplish this by regenerating the lattices between each iteration of extended Baum-Welch.  The resulting approximate MMI criterion is very close to the true MMI criterion at each iteration of extended Baum-Welch.  We observe that the resulting behavior of lattice-based MMI is much more benign, similar to what we observe with maximum likelihood estimation.  In particular nothing that could be labeled `over fitting' occurs. 

The remainder of this paper is organized as follows.  In Section~\ref{prelimiaries} we introduce the notation that we will be using, details concerning the approximate MMI criterion, and the particulars of our experimental set-up.  Section~\ref{experiments} describes and analyses the behavior of extended Baum-Welch using three different approximate MMI criteria.  We then wrap up the main body of the paper with a discussion in Section~\ref{discussion}.  Finally, we have also include three appendices that further analyze the experiment in Section~\ref{fixedLats}: they cover an alternate analysis of the behavior of the model parameters in Appendix~\ref{modelParameters}, the effect on our results of a parameter that controls the behavior of extended Baum-Welch in Appendix~\ref{E_experiments}, and preliminary results concerning MPE in Appendix~\ref{mpe}.

\subsection{Acknowledgments}

First of all, the core of the researcher described in this paper were obtained while the author was a researcher at VoiceSignal Technologies.  He is grateful to VoiceSignal for its support of this project.  The author is also extremely grateful to Nuance Communications for setting up what was essentially a sabbatical year for him in 2008.  This freedom helped him simplify and clarify the ideas presented below.  The author would also like to thank his colleagues Gunnar Evermann, Mike Newman, and Bob Roth for help with the lattice generation.  Lastly, the author is deeply indebted to Larry Gillick for his help with this and many other projects over the years, his sage advice, and his friendship.

\section{Preliminaries}\label{prelimiaries}

\subsection{The approximate MMI criterion}\label{notation}

Let \(X_1, X_2, \dots, X_n\) be a sequence of random, \(d\)-dimensional acoustic vectors, which we will abbreviate by \(X\) and let \(W\) be a random transcription taking values in \(\mathcal{W}\).    We let \((x,w)\) be the acoustics and the transcription of the training data that we will describe in Section~\ref{experimental_prelimiaries}.  We will denote the HMM-based probability model for \(X\), the acoustic model, by \(f_\theta\), where \(\theta\) are the model parameters that take values in \(\Theta\).  We will denote the probability distribution, or language model, for \(W\), simply by \(p\).\footnote{In practice there are really two language models in play in this paper.  The first is used exclusively on the training data for lattice generation, discriminative training, and recognition.  Following current standard practice (see \cite{schluter2} or \cite{povey2}), \(p\) is a relatively weak bigram language model estimated from the training transcriptions \(w\).  The second is used exclusively on our independent test set.  This distribution is a larger, bigram language model estimated from transcriptions disjoint from both the test and training transcriptions.}

The earlier versions of MMI used the conditional likelihood of the training data, \(p_\theta(w \mid x)\), as a model selection criterion which is given via Bayes' rule by  
\begin{equation}
p_\theta(w \mid x) = \frac{f_\theta(x \mid w) p(w ) }{\sum_{\textrm{w} \in \mathcal{W}} f_\theta(x \mid \textrm{w}) p(\textrm{w} )}.
\end{equation}
In principle the criterion that lattice-based MMI uses for model selection is a scaled version of \(p_\theta(w \mid x)\), which we denote by \(p_\theta(w \mid x; \kappa)\), that is defined by
\begin{equation} \label{scaledTheoretical}
p_\theta(w \mid x; \kappa) = \frac{f_\theta(x \mid w)^{\frac{1}{\kappa}} p(w ) }{\sum_{\textrm{w} \in \mathcal{W}} f_\theta(x \mid \textrm{w})^{\frac{1}{\kappa}} p(\textrm{w} )}. 
\end{equation}
The scale \(\kappa\), which is known as the \emph{language model scale}, is used in all practical recognition systems to balance the relative weights of the probabilities obtained from the language model and the acoustic model.  Since something analogous to the conditional likelihood \(p_\theta(w \mid x; \kappa)\) is used for hypothesis selection during recognition, it is also natural to use it as a model selection criterion.\footnote{This idea was first described in \cite{schluter}.}  In reality, however, lattice-based MMI actually uses an approximation to \(p_\theta(w \mid x; \kappa)\) for model selection.  This approximation, which has two aspects, is the result of efficiencies that extended Baum-Welch is able to make based on two properties of the lattices.

The first aspect of the approximation, which is common to all of the versions of MMI, involves the word level properties of the lattices and the sum in the denominator term of (\ref{scaledTheoretical}).  Instead of summing over all the possible transcriptions in \(\mathcal{W}\), which is impossible except in the simplest tasks, we restrict ourselves to a finite subset \(\mathcal{V}_{\theta_0} \subset \mathcal{W}\).  The subset \(\mathcal{V}_{\theta_0}\) is obtained by keeping all of the hypotheses with probability bigger than some small \(\epsilon > 0\) during a recognition over the training data using the seed acoustic models \(\theta_0\).  The correct transcript, \(w\), is also added to \(\mathcal{V}_{\theta_0}\).  We may think of \(\mathcal{V}_{\theta_0}\) as essentially\footnote{This is not precisely correct since a practical recognizer is unable to consider all possible \(\textrm{w} \in \mathcal{W}\) except on tasks like isolated digits.  Thus in general there will be some \(\bar{\textrm{w}} \in \mathcal{W} \setminus \mathcal{V}_{\theta_0}\) with \(p_{\theta_0}(\bar{\textrm{w}} \mid x; \kappa) \geq \epsilon\).  These \(\bar{\textrm{w}}\) are often called \emph{search errors}.} being defined by
\begin{equation} \label{definitionV}
\mathcal{V}_{\theta_0} = \{\textrm{w} \in \mathcal{W} : p_{\theta_0}(\textrm{w} \mid x; \kappa) \geq \epsilon\} \cup \{w\}.
\end{equation}
We use \(\mathcal{V}_{\theta_0}\) to construct an approximation to \(p_\theta(\cdot \mid x; \kappa)\) that we denote by \(p_\theta(\cdot \mid x; \mathcal{V}_{\theta_0})\) and that we define for any \(v \in \mathcal{V}_{\theta_0}\) by
\begin{equation} \label{firstSumApprox}
p_\theta(v \mid x ; \kappa, \mathcal{V}_{\theta_0}) = \frac{f_\theta(x \mid v)^{\frac{1}{\kappa}} p(v ) }{\sum_{\textrm{v} \in \mathcal{V}_{\theta_0}} f_\theta(x \mid \textrm{v})^{\frac{1}{\kappa}} p(\textrm{v} )}.
\end{equation}
The approximation \(p_\theta(\cdot \mid x ; \kappa, \mathcal{V}_{\theta_0})\) is also a probability mass function, however \textit{prima facie} not for \(W\), instead for \(V\), where \(V\) is the restriction of \(W\) from \(\mathcal{W}\) to \(\mathcal{V}_{\theta_0}\).  Of course, we can extend \(p_\theta(\cdot \mid x; \kappa, \mathcal{V}_{\theta_0})\) to be a probability mass function for \(W\) by setting it to zero on \(\mathcal{W} \setminus \mathcal{V}_{\theta_0}\).  How good the approximation
\begin{equation} \label{wordLevelApprox}
p_\theta(\cdot \mid x ; \kappa) \approx p_\theta(\cdot \mid x ; \kappa, \mathcal{V}_{\theta_0})
\end{equation}
is for arbitrary \(\theta\) depends on how large the missing probability density given by
\begin{equation} \label{missingWordMass}
\sum_{\text{w} \in \mathcal{W} \setminus \mathcal{V}_{\theta_0}} f_\theta(x \mid \textrm{w})^{\frac{1}{\kappa}} p(\textrm{w} )
\end{equation}
is.  At the seed models \(\theta_0\), provided that we have chosen the \(\epsilon\) in (\ref{definitionV}) small enough, the corresponding sum in (\ref{missingWordMass}) should be very small, so the approximation (\ref{wordLevelApprox}) should very close to an identity.  Hence for all \(\textrm{w} \in \mathcal{W}\)
\begin{equation*}
p_{\theta_0}(\textrm{w} \mid x ; \kappa) = p_{\theta_0}(\textrm{w} \mid x ; \kappa, \mathcal{V}_{\theta_0}).
\end{equation*}

The second aspect of the approximation involves the phone level properties of the lattices and the allowable state sequences in the definition of the hidden Markov model \(f_\theta\).  For any \(\textrm{w} \in \mathcal{W}\), let \(S_{\textrm{w}}\) denote the set of hidden state sequences, \(s\), that are compatible with transcription \(\textrm{w}\) and have the same number of frames as \(x\), namely \(n\).\footnote{For example, in this paper we will be using HMMs to model triphones.  So to construct the set \(S_{\textrm{w}}\) we first take all the phone-level pronunciations consistent with \(\textrm{w}\), then expand them to produce corresponding triphone level pronunciations, and finally enumerate the all of the state sequences with length \(n\) consistent with all of the possible triphone pronunciations.}  One facet of the model \(f_\theta(x \mid \textrm{w})\) is that we need to perform the following sum
\begin{equation} \label{hmmSum}
f_\theta(x \mid \textrm{w}) = \sum_{s \in S_{\textrm{w}}} f_\theta(x, s).
\end{equation}
Unfortunately, in the denominator of (\ref{firstSumApprox}) we need to compute \(f_\theta(x \mid \textrm{v})\) for every \(\textrm{v} \in \mathcal{V}_{\theta_0}\), which means that we need to efficiently evaluate sums analogous to (\ref{hmmSum}) for all the state sequences in \(S_{\textrm{v}}\), where \(\textrm{v}\) ranges over the large set \(\mathcal{V}_{\theta_0}\).  Lattice-based MMI gets around this problem by making use of so called \emph{phone-marked} word lattices.  Effectively this means that each transcription \(\textrm{w} \in \mathcal{V}_{\theta_0}\) has been force aligned using the seed models \(\theta_0\), but only the times at the phone boundaries -- these are the phone-marks -- are kept from the alignments.  The approximation involves restricting the sum in (\ref{hmmSum}) to a subset \(R_{\textrm{w}} \subset S_{\textrm{w}}\) that respects the phone-marks, in the sense that each phone's HMM is anchored at the start and end times given by the corresponding phone-marks.\footnote{This is accomplished in extended Baum-Welch by accumulating statistics using an phone-arc version of the forward-backward algorithm.  This is described in detail in \cite{povey2}.}  If we let
\begin{equation*}
\mathcal{R}_{\theta_0} = \{R_{\textrm{v}}\}_{\textrm{v} \in \mathcal{V}_{\theta_0}},
\end{equation*}
then given \(v \in \mathcal{V}_{\theta_0}\) we define an approximation to \(f_\theta(x \mid v)\), \(g_\theta( x \mid v ; \mathcal{R}_{\theta_0})\), by
\begin{equation*}
g_\theta(x \mid v ; \mathcal{R}_{\theta_0}) = \sum_{s \in R_{v}} f_\theta(x, s).
\end{equation*}

The combination of these two approximations results in the following definition\footnote{The notation that we are using for \(\mathcal{V}_{\theta_0}\) and \(\mathcal{R}_{\theta_0}\) is ambiguous because \(\mathcal{R}_{\theta_0}\) depends on \(\mathcal{V}_{\theta_0}\) and both depend on \(x\).  Fortunately, from now on, all these quantities will be appearing together, e.g. in \(p_\theta(v \mid x ; \kappa, \mathcal{V}_{\theta_0}, \mathcal{R}_{\theta_0})\), thus eliminating any ambiguity.}:
\begin{equation*}
p_\theta(v \mid x ; \kappa, \mathcal{V}_{\theta_0}, \mathcal{R}_{\theta_0}) = \frac{g_\theta(x \mid v; \mathcal{R}_{\theta_0})^{\frac{1}{\kappa}} p(v ) }{\sum_{\textrm{v} \in \mathcal{V}_{\theta_0}} g_\theta(x \mid \textrm{v}; \mathcal{R}_{\theta_0})^{\frac{1}{\kappa}} p(\textrm{v} )}.
\end{equation*}
The approximation \(p_\theta( \cdot \mid x ; \kappa, \mathcal{V}_{\theta_0}, \mathcal{R}_{\theta_0})\) to \(p_\theta( \cdot \mid x ; \kappa)\) is also a probability mass function for \(V\) which we extend to \(W\) by setting it to zero on \(\mathcal{W} \setminus \mathcal{V}_{\theta_0}\).  The approximate conditional likelihood of the training data, \(p_\theta(w \mid x ; \kappa, \mathcal{V}_{\theta_0}, \mathcal{R}_{\theta_0})\), is what lattice-based MMI actually uses as a model selection criterion.

If \(\mathcal{V}_{\theta_0}\) and \(\mathcal{R}_{\theta_0}\) are large enough, then for any \(\textrm{w} \in \mathcal{W}\) the approximation
\begin{equation} \label{probApprox}
p_\theta( \textrm{w} \mid x ; \kappa) \approx p_\theta( \textrm{w} \mid x ; \kappa, \mathcal{V}_{\theta_0}, \mathcal{R}_{\theta_0})
\end{equation}
should, in fact, be an equality at \(\theta_0\), namely
\begin{equation*}
p_{\theta_0}( \textrm{w} \mid x ; \kappa) = p_{\theta_0}( \textrm{w} \mid x ; \kappa, \mathcal{V}_{\theta_0}, \mathcal{R}_{\theta_0}).
\end{equation*}
This statement follows from construction.  How good the approximation in (\ref{probApprox}) remains as \(\theta\) moves away from \(\theta_0\) is one of the central questions in this paper.  Note that this approximation is only valid given the acoustic training data \(x\).

We shall be using the mle, which we denote by \(\theta_{mle}\), to generate the transcriptions and phone-marks in various combinations in our experiments.  Instead of writing \(\mathcal{V}_{\theta_{mle}}\) or \(\mathcal{R}_{\theta_{mle}}\) we shall simplify these by writing \(\mathcal{V}_{mle}\) or  \(\mathcal{R}_{mle}\) instead.  

We shall refer to the quantity
\begin{equation*}
\num(x, w ; \theta, \kappa, \mathcal{V}_{\theta_0}, \mathcal{R}_{\theta_0}) \equiv \log g_\theta(x \mid w; \mathcal{R}_{\theta_0})^{\frac{1}{\kappa}} p(w )
\end{equation*}
as the \emph{numerator log-likelihood} and the quantity
\begin{equation*}
\den(x; \theta, \kappa, \mathcal{V}_{\theta_0}, \mathcal{R}_{\theta_0}) \equiv \log \left(\sum_{\textrm{w} \in \mathcal{V}_{\theta_0}} g_\theta(x \mid \textrm{w}; \mathcal{R}_{\theta_0})^{\frac{1}{\kappa}} p(\textrm{w} )\right)
\end{equation*}
as the \emph{denominator log-likelihood}.

\subsection{Experimental preliminaries}\label{experimental_prelimiaries}

In this Section we give the details that all of our experiments share.  We chose to work on a standard Wall Street Journal (WSJ) task from the early 1990's (\cite{paul}, \cite{kubala}) because, by modern standards, it is small enough so that experimental turnaround is fast even with MMI, but it is large enough so that the results are believable.  This task is also self-contained, with nearly all of the materials necessary for training and testing available through the LDC, the exception being a dictionary for training and testing pronunciations.  We use pronunciations created at VoiceSignal Technologies (VST) using 39 non-silence phones. We use version 3.4 of the HTK toolkit to train and test our models.

We use the WSJ SI-284 set for acoustic model training.  This training set consists of material from 84 WSJ0 training speakers and from 200 WSJ1 training speakers.  It amounts to approximately 37000 training sentences and 66 hours of non-silence data.  Each session was recorded using two microphones; we use the primary channel recorded using a Sennheiser microphone.

The VST front-end that we use produces a 39 dimensional feature vector every 10 ms: 13 Mel-cepstral coefficients, including c0, plus their first and second differences.   The cepstral coefficients are mean normalized.  The data is down-sampled from 16 kHz to 8 kHz before the cepstral coefficients are computed.

We use very small, simple acoustic models to lower the computational load for MMI.  The acoustic models use word-internal triphones.  Except for silence, each triphone is modeled using a three state HMM without skipping.  For silence we follow the standard HTK practice that uses two models for silence: a three state tee-model and a single state short pause model; the short pause model is tied to middle state of the longer model; both models allow skipping.\footnote{See \cite{young} for details.} The resulting triphone states were then clustered using decision trees to 1500 tied states.  The output distribution for each tied state is a single, multivariate normal distribution with a diagonal covariance.

We report word error rate (WER) on two test sets.  Using the nomenclature of the time, these test sets use the 5k closed vocabulary and non-verbalized punctuation.  The first test set is the November 1992 ARPA evaluation 5k test set.  It has 330 sentences collected from 8 speakers.  The second test set is referred to as si\_dt\_05.odd in \cite{woodland1}.  It is a subset of the the WSJ1 5k development test set defined by first deleting sentences with out of vocabulary (OOV) words (relative to the 5k closed vocabulary) and then selecting every other sentence.  This results in 248 sentences collected from 10 speakers.  Together, these two test sets amount to about an hour of non-silence data.  We test using the standard 5k bigram language model created at Lincoln Labs for the 1992 ARPA evaluation.  The combined WER rate on these test sets using the models described above is 18\%.

When we refer to MMI training, we mean lattice-based extended Baum-Welch as described in \cite{povey2} or \cite{woodland2}.  We use HTK 3.4 to perform extended Baum-Welch with standard settings, e.g, \(E = 1\). We update the means and variances, but do not update the transition probabilities.  We use a VST tool and a relatively weak bigram language model estimated from the acoustic training sentences (we kept bigrams that had 8 or more examples) to generate word lattices on the training set.  We use HTK tools to create phone-marked numerator and denominator word lattices, the latter starting from the word lattices described above.  The language model scores in the phone-marked word lattices come from the weak bigram language model that was used to generate the word lattices.

For each feature we create a variance floor set to 1\% of the total variance of that feature in the training data.  These floors are respected during maximum likelihood and MMI training.

In all of our experiments, \(\kappa = 16\).  The HTK extended Baum-Welch software reports the per frame average of three quantities, namely, the numerator and denominator log-likelihoods, as well as their difference, i.e., the logarithm of the approximate MMI criterion.  When we report results we too shall report per frame averages of these quantities.

\subsection{A remark concerning \(\Theta\)} \label{thetaClosed}

Since we are only updating the model means and variances during MMI we shall think of the model parameter space \(\Theta\) as consisting of just the space of model means and variances.  Even with the small 1500 state unimodal models that we are using, \(\Theta\) is still quite large: on the order of \(10^5\) dimensional.  In general, if we define \(\mathcal{H}_{39} = \{\sigma^2 \in \mathbb{R}^{39}: \sigma_i^2 > 0 \textrm{ for } 1 \leq i \leq 39\}\), then each the variance and mean for state \(j\), \((\sigma_j^2, \mu_j)\), range over the product \(\mathcal{H}_{39} \times \mathbb{R}^{39}\), so
\begin{equation*}
\Theta = \prod_{j=1}^{1500} \left(\mathcal{H}_{39} \times \mathbb{R}^{39}\right)
\end{equation*}
which is an open set.  But as a practical matter, we are flooring the variances: if we let \(y \in \mathbb{R}\) be the variance floor, i.e. for each \(i\) with \( 1 \leq i \leq 39\) we have \(y_i > 0\), and we let \(\bar{\mathcal{H}}_{39} = \{\sigma^2 \in \mathbb{R}^{39}: \sigma_i^2 \geq 0 \textrm{ for } 1 \leq i \leq 39\}\) be the closure of \(\mathcal{H}_{39}\), then in actuality
\begin{equation*}
\Theta = \prod_{j=1}^{1500} \left((\bar{\mathcal{H}}_{39} + y) \times \mathbb{R}^{39}\right)
\end{equation*}
which is a closed set.

At several points in the paper it will be useful to refer to the distance between model parameters.  We will use the usual Euclidean distance on \(\Theta\), denoted \(\| \cdot \|\), to measure these distances.

\section{Experimental results}\label{experiments}

All of the experiments described in this section start from the mle, \(\theta_{mle}\), then construct a sequence of models parameters \((\theta_k)\) by iteratively applying the extended Baum-Welch algorithm 100 times.  The experiments differ in what lattices each iteration of extended Baum-Welch uses.  In the first experiment we shall follow the standard procedure in the literature that uses lattices generated the mle for each iteration of extended Baum-Welch.  This corresponds to using \(p_{\theta}( w \mid x ; \kappa, \mathcal{V}_{mle}, \mathcal{R}_{mle})\) as the model selection criterion during each iteration of extended Baum-Welch.  In the second experiment we shall regenerate the lattices between each iteration of extended Baum-Welch.  This corresponds to using \(p_\theta( w \mid x ; \kappa, \mathcal{V}_{\theta_k}, \mathcal{R}_{\theta_k})\) as the model selection criterion during iteration \(k+1\) of extended Baum-Welch.  In the third and final experiment we shall use the word lattices generated by the mle but we shall regenerate the phone-marks between each iteration of extended Baum-Welch.  This corresponds to using \(p_\theta( w \mid x ; \kappa, \mathcal{V}_{mle}, \mathcal{R}_{\theta_k})\) as the model selection criterion during iteration \(k+1\) of extended Baum-Welch.

\subsection{Fixed lattices} \label{fixedLats}

\begin{figure}
\centering
\includegraphics[width=1.0\textwidth]{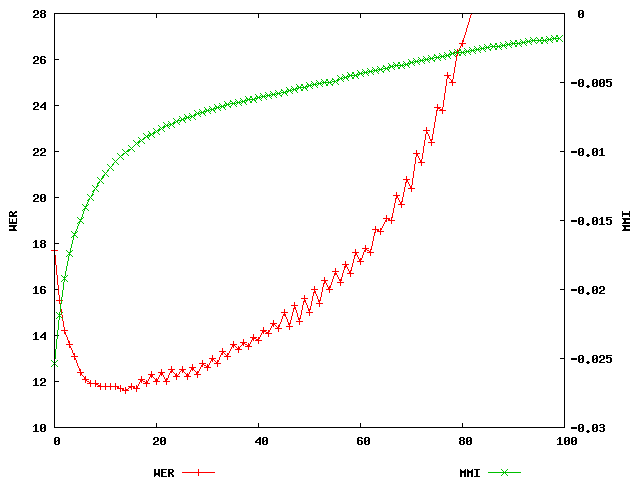}
\caption{The WER on an independent test set and the MMI criterion on the training data during 100 iterations of extended Baum-Welch.  The x-axis gives the extended Baum-Welch iteration, with x=0 being the mle.\label {longer_mmi_graph_repeat}} 
\end{figure}

In this experiment we generate the lattices once and for all using the mle, then run 100 iterations of extended Baum-Welch.  Figure \ref{longer_mmi_graph} in the introduction shows the results of this experiment, which we redisplay for the reader's convenience in Figure \ref{longer_mmi_graph_repeat}.  It is worth noting that  14 iterations of extended Baum-Welch reduces the WER of the mle, \(17.7 \%\),  to \(11.6 \%\).  This is a remarkable reduction in WER, which illustrates the utility of lattice-based MMI.  Unfortunately after 20 iterations the WER starts to steadily increase on the test data.  Indeed after 60 iterations the WER has exceeded that of the mle, and after 100 iterations the WER is \(42.4\%\) which is nearly 2.5 times worse than the WER of the mle.  This is in spite of the fact that the approximate MMI criterion is steadily increasing during these 100 iterations.  The approximate MMI criterion that is displayed in Figure \ref{longer_mmi_graph_repeat} is the sequence \((\log p_{\theta_k}( w \mid x ; \kappa, \mathcal{V}_{mle}, \mathcal{R}_{mle}))\).

Note that during the first 14 iterations the WER steadily decreases, in synchrony with the steady increase in the approximate MMI criterion. By construction the approximation
\begin{equation} \label {probApprox2}
p_{\theta}( w \mid x ; \kappa) \approx p_{\theta}( w \mid x ; \kappa, \mathcal{V}_{mle}, \mathcal{R}_{mle})
\end{equation}
should be quite reasonable for model parameters \(\theta\) near the mle, \(\theta_{mle}\).  Our hypothesis is that as extended Baum-Welch proceeds it gradually pushes the sequence \((\theta_k)\) away from the mle into a region where the approximation (\ref {probApprox2}) is no longer valid.

We supply further evidence for this hypothesis by running recognition on the training data using the sequence of model parameters produced by extended Baum-Welch, \((\theta_k)\), and two different recognition methods.  To perform recognition using the models \(\theta_k\) in the first method, which we refer to as `method A', we first generate the phone-marked training lattices using \(\theta_k\).  We then pick the best path through the phone-marked word lattices, which corresponds to choosing a transcription \(\textrm{w}^{*}_k \in \mathcal{W}\) according to the rule
\begin{equation*}
\textrm{w}^{*}_k = \arg \max_{\textrm{w} \in \mathcal{W}} p_{\theta_k}( \textrm{w} \mid x ; \kappa, \mathcal{V}_{\theta_k}, \mathcal{R}_{\theta_k}).
\end{equation*}
Since, as we have remarked in Section~\ref{notation}, the distributions \(p_{\theta_k}( \cdot \mid x ; \kappa, \mathcal{V}_{\theta_k}, \mathcal{R}_{\theta_k})\) and \(p_{\theta_k}( \cdot \mid x ; \kappa)\) should be essentially the same, this amounts to constructing a sequence of recognition transcriptions \((\textrm{w}^{*}_k)\) according to the rule
\begin{equation} \label{bigram_rule}
\textrm{w}^{*}_k = \arg \max_{\textrm{w} \in \mathcal{W}} p_{\theta_k}( \textrm{w} \mid x ; \kappa).
\end{equation}
In the second method, which we refer to as `method B', we first rescore the phone-marked word lattices generated by the mle -- the lattices that we used during extended Baum-Welch -- with the acoustic model with parameters \(\theta_k\), then pick pick the best path through the resulting lattices.  This amounts to constructing a sequence of recognition transcriptions \((v^{*}_k) \in \mathcal{V}_{mle}\) according to the rule
\begin{equation} \label{constraint_rule}
v^{*}_k = \arg \max_{v \in \mathcal{V}_{mle}} p_{\theta_k}( v \mid x ; \kappa, \mathcal{V}_{mle}, \mathcal{R}_{mle}).
\end{equation}

Table \ref {100iterationsEbwFixedLatWErTraining} displays the results of these recognitions as extended Baum-Welch proceeds.  The pattern of the WER of the recognition of the training data using method A is very similar to the pattern that we saw on the independent test data, but these patterns are very different from the pattern of the WER of the recognition of the training data using method B.  In particular, the sequence of WERs generated by method B steadily decreases, apparently headed towards \(0\%\) WER.

\begin{table}
\centering
\begin{tabular}{|r|c|c|}
\hline
 & \multicolumn{2}{|c|}{Method}\\ \cline{2-3}
iteration & A  & B\\
\hline
\hline
mle &     28.3 & 28.3\\
10  &     17.6 & 15.4\\
20  &     15.6 & 11.3\\
30  &     14.9 & 9.3\\
40  &     15.2 & 8.1\\
50  &     15.6 & 7.2\\
60  &     19.3 & 6.1\\
70  &     23.8 & 5.2\\
80  &     33.2 & 4.2\\
90  &     45.7 & 3.2\\
100 &     57.3 & 2.6\\
\hline
\hline
\end{tabular}
\caption{WER on training data during 100 iterations of extended Baum-Welch using two different recognition methods.\label {100iterationsEbwFixedLatWErTraining}}
\end{table}

As extended Baum-Welch proceeds the probability of the reference training transcription, \(p_{\theta_k}( w \mid x ; \kappa, \mathcal{V}_{mle}, \mathcal{R}_{mle})\), steadily increases, while the sequence of transcriptions recognized by the rule (\ref{constraint_rule}), \((v^{*}_k)\), has steadily decreasing WER.  Since the WER is steadily decreasing, the sequence of recognized transcriptions, \((v^{*}_k)\), is moving closer to the reference transcription \(w\).  It follows that extended Baum-Welch is making the reference transcription, or a transcription very close to it, the most likely transcription under the rule (\ref{constraint_rule}) which uses the approximation \(p_{\theta_k}( \cdot \mid x ; \kappa, \mathcal{V}_{mle}, \mathcal{R}_{mle})\) for \(k \gg 0\).  In contrast, the WER of the sequence of transcriptions recognized via the rule (\ref{bigram_rule}), \((\textrm{w}^{*}_k)\) is steading increasing after iteration 30 which means that this sequence of recognized transcriptions must be steadily moving further away from the reference transcription \(w\).  Since the probability distributions \(p_{\theta_k}( \cdot \mid x ; \kappa, \mathcal{V}_{mle}, \mathcal{R}_{mle})\) and \(p_{\theta_k}( \cdot \mid x ; \kappa)\) are making drastically different choices under their respective recognition rules for large \(k\), it follows these probability distributions must be very different after many iterations of extended Baum-Welch.

\begin{figure}
\centering
\includegraphics[width=.75\textwidth]{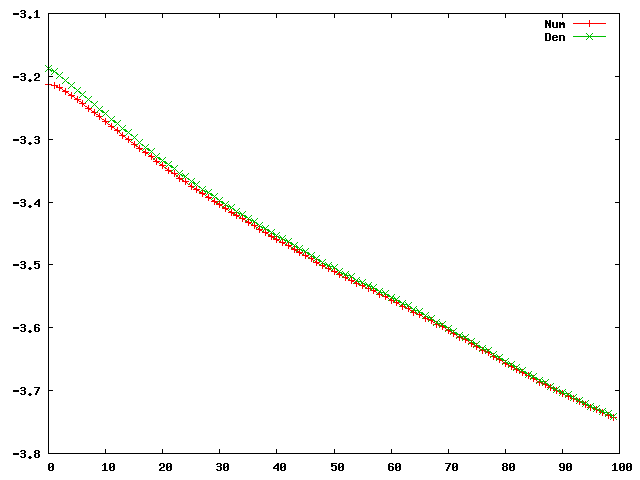}
\caption{Numerator and denominator log-likelihoods during 100 iterations of extended Baum-Welch.\label{numdenebw}}
\end{figure}

Why does the approximation in (\ref{probApprox}) break down for \(k \gg 0\)?  To answer this, we examine the sequence of numerator log-likelihoods
\begin{equation*}
(\num(x, w ; \theta_k, \kappa, \mathcal{V}_{mle}, \mathcal{R}_{mle}))
\end{equation*}
and the sequence of denominator log-likelihoods
\begin{equation*}
(\den(x, w ; \theta_k, \kappa, \mathcal{V}_{mle}, \mathcal{R}_{mle}))
\end{equation*}
over the 100 iterations in Figure \ref {numdenebw}.  Since the sequence of numerator log-likelihoods is steadily decreasing, the sequence of parameters \((\theta_k)\) must be steadily moving away from \(\theta_{mle}\).  The sequence of denominator log-likelihoods is not only steadily decreasing but is also moving closer to the the sequence of numerator log-likelihoods.  From the definitions of the numerator and denominator log-likelihoods, it follows that 
\begin{equation} \label{shakyApprox}
g_{\theta_k}(x \mid w; \mathcal{R}_{mle})^{\frac{1}{\kappa}} p(w ) \gg g_{\theta_k}(x \mid \textrm{w}; \mathcal{R}_{mle})^{\frac{1}{\kappa}} p(\textrm{w} )
\end{equation}
for large \(k\) and every transcription \(\textrm{w} \in \mathcal{V}_{mle}\) different from the reference transcription \(w\).  This property of \(g_{\theta_k}\) and \(\mathcal{V}_{mle}\) is not shared by \(f_{\theta_k}\) and \(\mathcal{W}\).  We have already shown that for large \(k\) the transcript \(\textrm{w}^{*}_k \in \mathcal{W}\) recognized by rule (\ref{bigram_rule}) is very different from \(w\), hence the the analog of (\ref{shakyApprox}) does not hold for \(f_{\theta_k}\) since \(\textrm{w}^{*}_k \not = w\) and
\begin{equation} \label{trueApprox}
 f_{\theta_k}(x \mid \textrm{w}^{*}_k)^{\frac{1}{\kappa}} p(\textrm{w}^{*}_k) > f_{\theta_k}(x \mid w)^{\frac{1}{\kappa}} p(w ).
\end{equation}
We conclude that the reason that probability distributions \(p_{\theta_k}( \cdot \mid x ; \kappa, \mathcal{V}_{mle}, \mathcal{R}_{mle})\) and \(p_{\theta_k}( \cdot \mid x ; \kappa)\) are different is because by restricting the relevant sums in the definition of \(p_{\theta_k}( \cdot \mid x ; \kappa, \mathcal{V}_{mle}, \mathcal{R}_{mle})\) to \(\mathcal{R}_{mle}\) and \(\mathcal{V}_{mle}\) leads to ignoring state sequences in \(\mathcal{S} \setminus \mathcal{R}_{mle}\) or transcriptions in \(\mathcal{W} \setminus \mathcal{V}_{mle}\) that have zero probability under \(\theta_{mle}\) but non-trivial probability under \(\theta_k\).

However the situation is actually much worse than this analysis suggests.  To see this we turn to investigating the convergence properties of \((\theta_k)\).  As we noted in the introduction, Figure \ref{longer_mmi_graph_repeat} shows that the the sequence of model parameters \((\theta_k)\) cannot have converged during these 100 iterations, since the sequence \((\log p_{\theta_k}( w \mid x ; \kappa, \mathcal{V}_{mle}, \mathcal{R}_{mle}))\) has clearly not converged.  If the trend exhibited in Figure \ref {numdenebw} continues, then
\begin{equation} \label{ill-posed:root}
\lim_{k \to \infty} \num(x, w ; \theta_k, \kappa, \mathcal{V}_{mle}, \mathcal{R}_{mle}) = \lim_{k \to \infty} \den(x, w ; \theta_k, \kappa, \mathcal{V}_{mle}, \mathcal{R}_{mle}) = -\infty.
\end{equation}
It (\ref{ill-posed:root}) were true, then the continuity of \(\num(x, w ; \theta, \kappa, \mathcal{V}_{mle}, \mathcal{R}_{mle})\) as a function of \(\theta\) would show that not only is the sequence \((\theta_k)\) not convergent, but that it does not remain within any given compact set.  Furthermore, as we remarked in Section~\ref{thetaClosed}, because we are flooring the variances \(\Theta\) is in fact closed, so it would follow that the sequence \((\theta_k)\) must be unbounded.   More precisely, the sequence of model parameters \((\theta_k)\) created by 100 iterations of extended Baum-Welch is consistent with the following conjecture:\footnote{The observed conditional log-likelihood of the training data does change dramatically after 100 iterations of extended Baum-Welch, in fact  decreasing in absolute value by a factor of 14, which means that the relationship between the conditional likelihoods before and after 100 iterations of extended Baum-Welch is given by is
\begin{equation*}
\log p_{\theta_{100}}( w \mid x ; \kappa, \mathcal{V}_{mle}, \mathcal{R}_{mle}) \approx \left(\log p_{\theta_{mle}}( w \mid x ; \kappa, \mathcal{V}_{mle}, \mathcal{R}_{mle}) )\right)^{\frac{1}{14}}.
\end{equation*}
Thus the probability mass is considerably more concentrated on \(w\) under \(\theta_{100}\) than it was to begin with under \(\theta_{mle}\).}
\begin{conjecture}\label{mainconjecture}
Let \((\theta_k) \in \Theta\) be the sequence of model parameters defined inductively by \(\theta_0 = \theta_{mle}\), and \(\theta_{k+1}\) is obtained from \(\theta_k\) by using extended Baum-Welch with \(p_{\theta}( w \mid x ; \kappa, \mathcal{V}_{mle}, \mathcal{R}_{mle})\) as the model selection criterion.  Then
\begin{equation} \label{goestoinfty}
\lim_{k \to \infty} \|\theta_k\| = \infty
\end{equation}
and
\begin{equation} \label{supequals0}
\lim_{k \to \infty} \log p_{\theta_k}( w \mid x ; \kappa, \mathcal{V}_{mle}, \mathcal{R}_{mle}) = 0.
\end{equation}
\end{conjecture}

Conjecture \ref{mainconjecture} suggests that the approximate MMI criterion, \(p_{\theta}( w \mid x ; \kappa, \mathcal{V}_{mle}, \mathcal{R}_{mle})\), is a terrible choice for a model selection criterion.  This is because the problem
\begin{equation} \label{illPosed}
\hat{\theta} = \arg \max_{\theta \in \Theta} p_{\theta}( w \mid x ; \kappa, \mathcal{V}_{mle}, \mathcal{R}_{mle})
\end{equation}
admits an absurd solution, namely a point at infinity with
\begin{equation*}
 p_{\hat{\theta}}( w \mid x ; \kappa, \mathcal{V}_{mle}, \mathcal{R}_{mle}) = 1.
\end{equation*}
In other words, this would mean that the model selection problem in (\ref{illPosed}) is ill-posed.  On the other hand, the algorithm, extended Baum-Welch, appears to be blameless since it is doing what the apparently ill-posed problem requires it to do. 

One possible mechanism that extended Baum-Welch could use to drive the numerator and denominator log-likelihoods to big -- but not infinite -- negative values is to simply push a large number of the model variances to the variance floor.  We rule this mechanism out by observing that none of the mle model parameters have floored variances, while after 100 iterations of extended Baum-Welch only 10 out of the approximately 59,000 variances have been floored.  In Appendix~\ref{modelParameters} we give a more detailed analysis of what happens to the model parameters after 100 passes of extended Baum-Welch.  In particular, we show that there is an significant expansion of the space that the means occupy.

Finally, as we noted in Section~\ref{experimental_prelimiaries}, the extended Baum-Welch parameter \(E\) was set to 1 in this experiment.  It is natural to ask if larger values of \(E\) would result in different results.  In particular, we might speculate that it is possible to choose \(E\) large enough to guarantee convergence of the resulting sequence of model parameters \((\theta_k)\).  In Appendix~\ref{E_experiments} we examine this question.  There we find that increasing \(E\) only slows the behavior that we have observed in this section, with the fundamental problem of parameter divergence remaining with larger choices for \(E\).  However, this result should not be surprising since it is consistent with the results of this section: the parameter divergence, among the other deficiencies of lattice-based MMI, is due to properties of the approximate MMI criterion and not due to properties of extended Baum-Welch. 

\subsection{Regenerated lattices: word and phone-marks} \label{regenWordPhoneLats}

\begin{figure}
\centering
\includegraphics[width=1.0\textwidth]{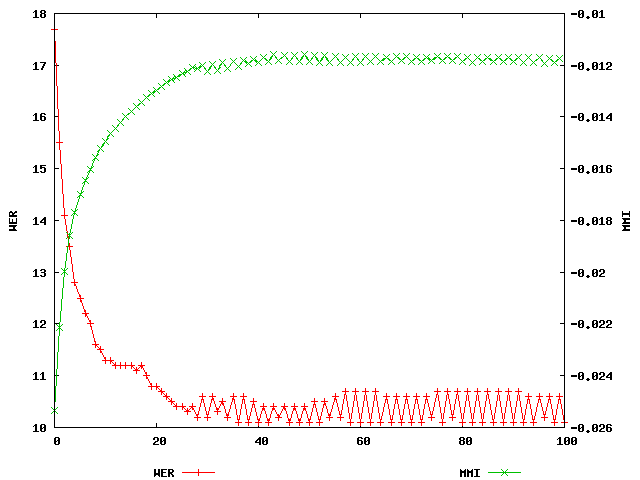}
\caption{MMI criterion on the training set during 100 iterations of extended Baum-Welch with lattice regeneration. The x-axis gives the extended Baum-Welch iteration, with x=0 being the mle.\label{mmi_regen_wer_mmi}}
\end{figure}

In this experiment we regenerate the phone-marked word lattices between each iteration of extended Baum-Welch.  Thus at iteration \(k+1\) we are using \(p_{\theta}( w \mid x ; \kappa, \mathcal{V}_{\theta_k}, \mathcal{R}_{\theta_k})\) as the model selection criterion when we estimate the parameters \(\theta_{k+1}\).  Starting from the mle, we run extended Baum-Welch 100 times.  We observe the following:
\begin{itemize}
\item[(1)] The logarithm of the approximate MMI criterion, \(\log p_{\theta_k}( w \mid x ; \kappa, \mathcal{V}_{\theta_k}, \mathcal{R}_{\theta_k})\), increases for the first 30 iterations and then starts to oscillate.  It reaches its peak value after 50 iterations and then continues its oscillation.  See Figure \ref {mmi_regen_wer_mmi}.
\item[(2)] The WER on the test set steadily decreases for the first 25 iterations, then continues to decrease but with oscillation.  After 36 iterations the WER reaches its minimum value, \(10.1\%\),  and then oscillates from \(10.1\%\) to various values ranging from \(10.4\%\) to \(10.9\%\).  See Figure \ref {mmi_regen_wer_mmi}.
\item[(3)] When we when measure the WER on the training data to three significant figures it steadily decreases and in fact appears to converge to \(12.3\%\).  See Table \ref{denlatrecmmieachtime}.  However, when we measure this WER to four significant digits, there is a small oscillation of \(\pm 0.05\%\).  
\end{itemize}
Using a matched pairs test (\cite{gillick}), the minimum WER on the test set in this experiment, \(10.1\%\), is significantly better than that of the previous experiment, \(11.6\%\), at a confidence level \( < 0.001\) (the smallest non-zero p-value that we can detect with our software is 0.001).  But the difference in the relative reduction in the WERs \(17.7 \to 11.6 = 34\%\) versus \(17.7 \to 10.1 = 43\%\) are similar.  Perhaps the improvement that we are seeing is artifact of the inherent noise in WERs obtained using simple models on small test sets. 

\begin{table}
\centering
\begin{tabular}{|r|c|}
\hline
iteration & WER\\
\hline
\hline
mle & 28.3 \\
10 &  17.4 \\
20 &  14.6 \\
30 &  13.3 \\
40 &  12.7 \\
50 &  12.5 \\
60 &  12.5 \\
70 &  12.4 \\
80 &  12.3 \\
90 &  12.3 \\
100 & 12.3 \\
\hline
\hline
\end{tabular}
\caption{WER on the training set during 100 iterations of extended Baum-Welch.\label {denlatrecmmieachtime}}
\end{table}

For model parameters \(\theta\) near \(\theta_k\) the approximation
\begin{equation} \label{approxRegenMmi}
p_\theta(w \mid x; \kappa) \approx p_\theta(w \mid x ; \kappa, \mathcal{V}_{\theta_k}, \mathcal{R}_{\theta_k})
\end{equation}
should be very good.  So by regenerating the phone-marked word lattices between each iteration of extended Baum-Welch, we are effectively using something very close to \(p_\theta(w \mid x; \kappa)\) as our model selection criterion for each iteration of extended Baum-Welch.  Figure \ref {mmi_regen_wer_mmi} shows that while the sequence \((p_{\theta_k}(w \mid x; \kappa))\) is oscillating for \(k\) large, it is not being driven to 1.  In other words, the model selection criterion \(p_\theta(w \mid x; \kappa)\) is not exhibiting the pathological behavior that \(p_\theta(w \mid x; \kappa, \mathcal{V}_{\theta_{mle}}, \mathcal{R}_{\theta_{mle}})\) exhibits.  Also, the sequences of word error rates on the independent test set and the training data behave similarly as extended Baum-Welch proceeds and are consistent with the sequence \(((p_{\theta_k}(w \mid x; \kappa))\) .  In particular we do not any phenomena that might be labeled `over training' or `over fitting'. 

Finally, even though we are seeing much more benign behavior than in the previous experiment with fixed lattices, neither the approximate MMI criterion has converged after 100 iterations of extended Baum-Welch, nor has the sequence of underlying parameters, \((\theta_k)\), converged.  However, in contrast to the previous experiment with fixed lattices, Figure \ref {mmi_regen_wer_mmi} suggests that in the limit the sequence \((\theta_k)\) is orbiting a compact limit set.

\subsection{Regenerated lattices: only phone-marks} \label{regenPhoneLatsOnly}

\begin{figure}
\centering
\includegraphics[width=1.0\textwidth]{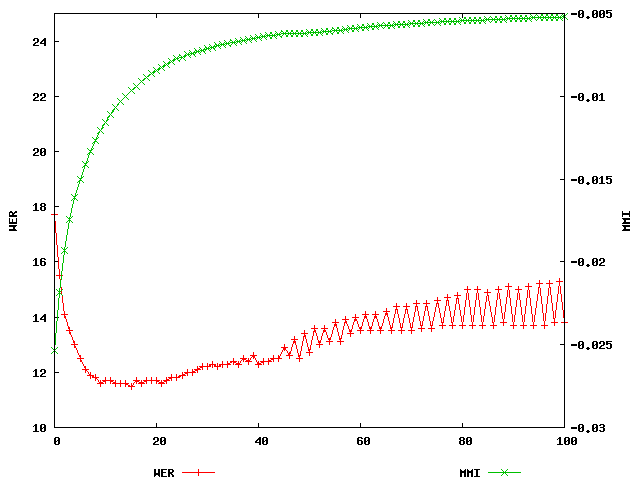}
\caption{MMI criterion on the training data during 100 iterations of extended Baum-Welch with lattice regeneration. The x-axis gives the extended Baum-Welch iteration, with x=0 being the mle.\label{phonemark_wer_mmi_100}}
\end{figure}

In this experiment we generate the word lattices once and for all using the mle, but we regenerate the phone-marks in the lattices between each iteration of extended Baum-Welch.  Thus at iteration \(k+1\) we are using \(p_{\theta}( w \mid x ; \kappa, \mathcal{V}_{mle}, \mathcal{R}_{\theta_k})\) as the model selection criterion when we estimate the parameters \(\theta_{k+1}\).  Starting from the mle, we run extended Baum-Welch 100 times.  The results we observe are a less extreme version of what we saw in Section~\ref{fixedLats}, namely the approximate MMI criterion steadily increases, while the WER on the independent test set steadily decreases until it reaches its minimum of \(11.5\%\) at iteration 15, levels out between iterations 16 and 22 where it finally then starts to gradually increase.  Significant oscillations in the WER begin after iteration 47.  See Figure \ref {phonemark_wer_mmi_100}.  So just as in the experiment with fixed lattices in Section~\ref{fixedLats}, we observe a disconnect between the steadily increasing approximate MMI criterion and the increasing test set WER.

Following the example set forward in Section~\ref{fixedLats}, we run recognition on the training data using two different recognition methods.  The first method is exactly the same as method A in Section~\ref{fixedLats}, namely we use the distribution \(p_{\theta_k}( \cdot \mid x ; \kappa)\) to select our recognition transcription at iteration \(k\) via the rule (\ref{bigram_rule}).  The second method, which we refer to as `method C' is analogous to method B in Section~\ref{fixedLats}, except that here we use the distribution \(p_{\theta_k}( \cdot \mid x ; \kappa, \mathcal{V}_{mle}, \mathcal{R}_{\theta_k})\) to select the recognition transcription instead of the distribution \(p_{\theta_k}( \cdot \mid x ; \kappa, \mathcal{V}_{mle}, \mathcal{R}_{mle})\) in the rule (\ref{constraint_rule}). The resulting recognition results are displayed in Table \ref{phoneMarkEachTimeComp}.  Notice that the sequence of WERs are both slowing decreasing over this range.  An argument analogous to what we presented in Section~\ref{fixedLats} shows that the probability distributions \(p_{\theta_k}(\cdot  \mid x ; \kappa)\) and \(p_{\theta_k}( \cdot \mid x ; \kappa, \mathcal{V}_{mle}, \mathcal{R}_{\theta_k})\) must be very different for \(k \gg 0\).  The only possible explanation for why these distributions are different is: by restricting the relevant sum in the definition of \(p_{\theta_k}( \cdot \mid x ; \kappa, \mathcal{V}_{mle}, \mathcal{R}_{k})\) to \(\mathcal{V}_{mle}\) leads to ignoring transcriptions in \(\mathcal{W} \setminus \mathcal{V}_{mle}\) that have zero probability under \(\theta_{mle}\) but non-trivial probability under \(\theta_k\).

\begin{table}
\centering
\begin{tabular}{|r|c|c|}
\hline
 & \multicolumn{2}{|c|}{Method}\\ \cline{2-3}
iteration & A & C\\
\hline
\hline
mle &     28.3 & 28.3\\
10  &     17.3 & 15.1\\
20  &     14.7 & 11.0\\
30  &     13.3 & 9.1\\
40  &     12.8 & 8.1\\
50  &     12.7 & 7.7\\
60  &     12.6 & 7.3\\
70  &     12.4 & 7.0\\
80  &     12.4 & 6.8\\
90  &     12.3 & 6.6\\
100 &     12.2 & 6.4\\
\hline
\hline
\end{tabular}
\caption{WER on training data during 100 iterations of extended Baum-Welch using two different recognition methods.\label {phoneMarkEachTimeComp}}
\end{table}

\begin{figure}
\centering
\includegraphics[width=1.0\textwidth]{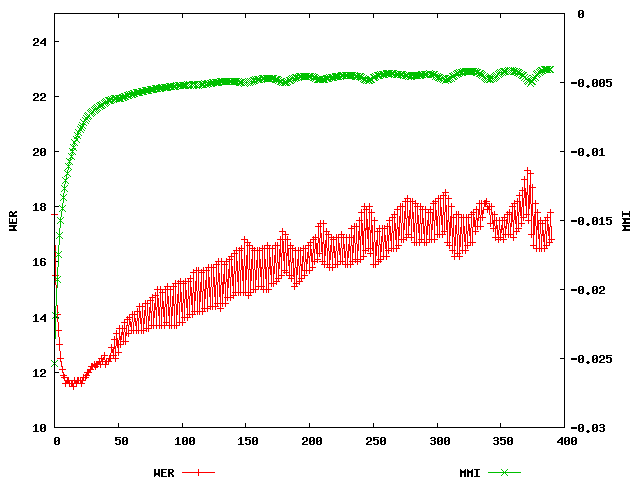}
\caption{WER on the test set during 100 iterations of extended Baum-Welch with lattice regeneration. The x-axis gives the extended Baum-Welch iteration, with x=0 being the mle.\label{phonemark_wer_mmi_380}}
\end{figure}

While the WER on the independent test set is increasing in Figure \ref {phonemark_wer_mmi_100} after iteration 22, the increase is very gradual and the oscillation in the WER is large.  We decided to run 290 more iterations of extended Baum-Welch to see if the minimum value in the oscillating WER ever exceeds that of the mle, namely \(17.7\%\).  Figure \ref{phonemark_wer_mmi_380} displays the results of the resulting 390 iterations from the beginning.  At iterations 337-341 the oscillation in the WER temporarily stops, with the WER at \(18.2\%\).

\begin{figure}
\centering
\includegraphics[width=1.0\textwidth]{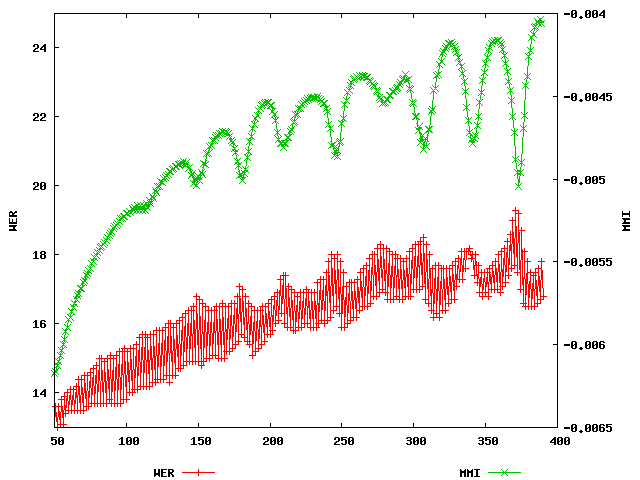}
\caption{WER on the test set during 100 iterations of extended Baum-Welch with lattice regeneration. The x-axis gives the extended Baum-Welch iteration, with x=0 being the mle.\label{phonemark_wer_mmi_50_380}}
\end{figure}

In Figure \ref{phonemark_wer_mmi_380} we see that the approximate MMI criterion begins to oscillate after about iteration 100.  The test set WER also begins a secondary oscillation at about that iteration.  This is easier to see in Figure \ref {phonemark_wer_mmi_50_380}.  Note that even though the overall trend in  approximate MMI criterion and test set WER is upward until about iteration 300, which means that there is a disconnect in the approximate MMI criterion from the test set WER, the oscillations in the approximate MMI criterion are connected to the secondary oscillations in the test set WER.  For example at iteration 371 the approximate MMI criterion is at the bottom of a steep valley, while the test set WER is at the top of a peak (\(19.2\%\)).

Finally we note that while the overall trend in both the approximate MMI criterion and the WER appear to be converging -- both trends start to level out at around iteration 300 -- the amplitudes of the secondary oscillations appear to be increasing.  Thus, even after nearly 400 iterations, the sequence of model parameters is not close to converging.  Also, because the secondary oscillations are increasing in amplitude, it is possible that the sequence of model parameters is heading to a point at infinity, but more extended Baum-Welch iterations and further analysis would be required to settle this point.

\subsection{Further analysis of the three experiments} \label{extraAnalysis}

In this section we present an analysis that shows that if we use \(p_{\theta}( w \mid x ; \kappa, \mathcal{V}_{mle}, \mathcal{R}_{mle})\) as our approximation to the MMI criterion, then
\begin{itemize}
\item[(1)]  For the first 9 iterations of extended Baum-Welch, this approximate MMI criterion consistently overestimates the value of its choice of model parameter relative to the value that the true MMI criterion would place on it.  We express this mathematically, in the range \(0 \leq k \leq 8\), by:
\begin{equation*}
p_{\theta_{k+1}}( w \mid x ; \kappa) < p_{\theta_{k+1}}( w \mid x ; \kappa, \mathcal{V}_{mle}, \mathcal{R}_{mle}).
\end{equation*}
\item[(2)] This overestimation is solely due to ignoring alternative transcriptions that the mle placed zero probability on.
\item[(3)] Significant errors in this approximation due to ignoring alternative state sequences, that the mle placed zero probability on, do not occur until after iteration 27. 
\end{itemize} 

\begin{figure}
\centering
\includegraphics[width=1.0\textwidth]{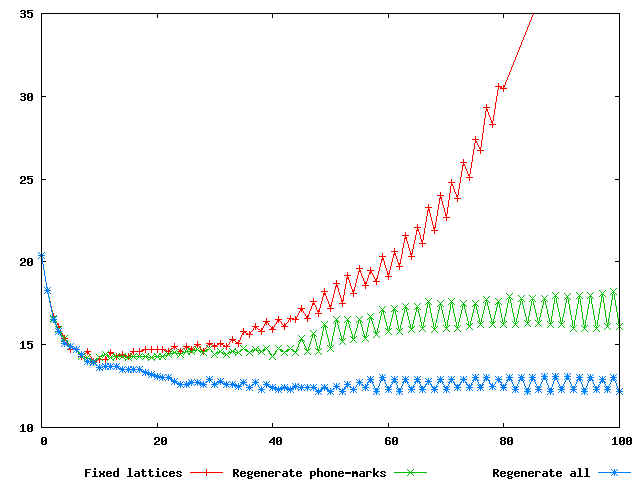}
\caption{Comparison of WERs on the independent test set during 100 iterations of extended Baum-Welch with the three types of input lattices. The x-axis gives the extended Baum-Welch iteration, with x=0 being the mle.\label{comparison_test_wer}}
\end{figure}

Figure \ref{comparison_test_wer} compares the WERs on the independent test set as extended Baum-Welch proceeds using the three approximate MMI criteria:  \(p_{\theta}( w \mid x ; \kappa, \mathcal{V}_{mle}, \mathcal{R}_{mle})\) which is labeled `Fixed lattices', \(p_{\theta}( w \mid x ; \kappa, \mathcal{V}_{mle}, \mathcal{R}_{k})\) which is labeled `Regenerate phone-marks', and \(p_{\theta}( w \mid x ; \kappa, \mathcal{V}_{k}, \mathcal{R}_{k})\) which is labeled `Regenerate all'.  These WERs are all essentially the same until iteration 10, when the `Regenerate all' WERs separate from the other two.  We infer from this that the corresponding sequences of model parameters follow the same pattern: all three sequences of model parameters are essentially the same until iteration 10, when the sequence created using the model selection criterion  \(p_{\theta}( w \mid x ; \kappa, \mathcal{V}_{k}, \mathcal{R}_{k})\) separates from the other two sequences.

\begin{figure}
\centering
\includegraphics[width=1.0\textwidth]{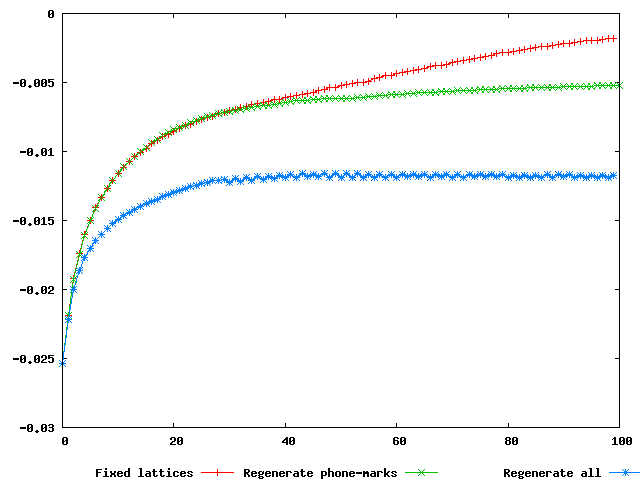}
\caption{Comparison of the approximate MMI criterion during 100 iterations of extended Baum-Welch with the three types of input lattices. The x-axis gives the extended Baum-Welch iteration, with x=0 being the mle.\label{comparison_mmi_criterion}}
\end{figure}

Similarly, Figure \ref{comparison_mmi_criterion} compares the values of the three approximate MMI criteria as extended Baum-Welch proceeds.  For the first nine iterations all three sequences of model parameters are the same, but in the range \(0 \leq k \leq 8\)
\begin{equation} \label{wordInequality}
p_{\theta_{k+1}}( w \mid x ; \kappa, \mathcal{V}_{k}, \mathcal{R}_{k}) < p_{\theta_{k+1}}( w \mid x ; \kappa, \mathcal{V}_{mle}, \mathcal{R}_{k}) = p_{\theta_{k+1}}( w \mid x ; \kappa, \mathcal{V}_{mle}, \mathcal{R}_{mle}).
\end{equation}
The definitions show that inequality in (\ref{wordInequality}) must be due to alternate training transcriptions \(\textrm{w} \in \mathcal{W}\setminus\mathcal{V}_{mle}\) satisfying
\begin{equation*}
g_{\theta_{mle}}(x \mid \textrm{w}; \mathcal{R}_{mle}) = 0
\end{equation*}
and, for \(k\) with \(0 \leq k \leq 8\),
\begin{equation*}
g_{\theta_{k+1}}(x \mid \textrm{w}; \mathcal{R}_{\theta_{k}}) = g_{\theta_{k+1}}(x \mid \textrm{w}; \mathcal{R}_{mle}) > 0.
\end{equation*}
Recall that the approximate MMI criterion \(p_{\theta_{k+1}}( w \mid x ; \kappa, \mathcal{V}_{k}, \mathcal{R}_{k})\) is essentially the same as \(p_{\theta_{k+1}}( w \mid x ; \kappa)\).  However, we have shown that the approximate MMI criteria \(p_{\theta_{k+1}}( w \mid x ; \kappa, \mathcal{V}_{mle}, \mathcal{R}_{k})\) and \(p_{\theta_{k+1}}( w \mid x ; \kappa, \mathcal{V}_{mle}, \mathcal{R}_{mle})\) are overly optimistic relative to \(p_{\theta_{k+1}}( w \mid x ; \kappa)\) solely because the corresponding distributions place zero probability on transcriptions that are not in \(\mathcal{V}_{mle}\) but have non-zero probability under \(p_{\theta_{k+1}}( \cdot \mid x ; \kappa)\).

Figures \ref{comparison_test_wer} and \ref{comparison_mmi_criterion} show that the test set WERs and the approximate MMI criteria are very similar for the first 27 iterations when we use the approximations \(p_{\theta_{k+1}}( w \mid x ; \kappa, \mathcal{V}_{mle}, \mathcal{R}_{mle})\) and \(p_{\theta_{k+1}}( w \mid x ; \kappa, \mathcal{V}_{mle}, \mathcal{R}_{\theta_k})\).  Thus a similar argument shows that it is not until after iteration 28 that ignoring state sequences in \(\mathcal{S}\setminus\mathcal{R}_{mle}\) starts to degrade the approximation in \(p_{\theta_{k+1}}( w \mid x ; \kappa, \mathcal{V}_{mle}, \mathcal{R}_{mle})\). 

Finally, Figures \ref{comparison_test_wer} and \ref{comparison_mmi_criterion} also make clear how important the regenerating the phone-marks are to good asymptotic behavior for lattice-based MMI.

\section{Discussion} \label{discussion}

We showed in Section~\ref{regenWordPhoneLats} that if we iteratively estimate model parameters using extended Baum-Welch and an approximate MMI criterion -- \(p_{\theta}( w \mid x ; \kappa, \mathcal{V}_{\theta_k}, \mathcal{R}_{\theta_k})\) at iteration \(k+1\) -- that is a consistently good approximation to \(p_{\theta}( w \mid x ; \kappa)\), then the resulting behavior is nearly ideal. The WERs on the training and test data decline while the approximate MMI criterion rises.  In fact the minimum WER on the test set is lower -- albeit by a small amount -- than when we use the standard approximate MMI criterion.  The sequences of WERs and approximate MMI criteria do not appear to actually converge, but instead oscillate within a small range of values, which suggests that the underlying parameters will asymptotically orbit a compact limit set.  In particular, there is no evidence that anything like `over fitting' is occurring.  On the contrary, this suggests that parameter estimation using the ideal MMI criterion, \(p_{\theta}( w \mid x ; \kappa)\), and the extended Baum-Welch algorithm should have properties that are nearly as good as maximum likelihood estimation using the Baum-Welch algorithm.

In contrast, we showed in Section~\ref{fixedLats} that the properties of the standard approximation to the MMI criterion, namely \(p_{\theta}( w \mid x ; \kappa, \mathcal{V}_{mle}, \mathcal{R}_{mle})\), make it unsuitable as a model selection criterion.  First of all, when \(\theta = \theta_{mle}\) the approximation
\begin{equation} \label{mmiCritApprox}
 p_{\theta}( w \mid x ; \kappa) \approx  p_{\theta}( w \mid x ; \kappa, \mathcal{V}_{mle}, \mathcal{R}_{mle})
\end{equation}
is essentially an equality and this approximation remains good while \(\theta\) is close to \(\theta_{mle}\).  However, we showed that as \(\theta\) moves away from \(\theta_{mle}\) the two distributions \(p_{\theta}( \cdot \mid x ; \kappa)\) and \(p_{\theta}( \cdot \mid x ; \kappa, \mathcal{V}_{mle}, \mathcal{R}_{mle})\) become very different. This is because, by construction, \(p_{\theta}( \cdot \mid x ; \kappa, \mathcal{V}_{mle}, \mathcal{R}_{mle})\) ignores transcriptions in \(\mathcal{W} \setminus \mathcal{V}_{mle}\) and state sequences in \(\mathcal{S} \setminus \mathcal{R}_{mle}\) that have zero probability under our model at \(\theta_{mle}\).  As \(\theta\) moves away from \(\theta_{mle}\) these ignored transcriptions and state sequences start to accumulate non-trivial probability mass, leading this approximate MMI criterion to consistently overestimate \(p_{\theta}( w \mid x ; \kappa)\).  Interestingly, we showed in Section~\ref{regenPhoneLatsOnly} and Section~\ref{extraAnalysis} that the ignored state sequences appear to contribute just as much as the ignored transcriptions do to the discrepancy between \(p_{\theta}( w \mid x ; \kappa, \mathcal{V}_{mle}, \mathcal{R}_{mle})\) and \(p_{\theta}( w \mid x ; \kappa)\) when \(\theta\) is far from \(\theta_{mle}\).  Second of all, and to make matters worse, we also showed that the approximate MMI criterion \(p_{\theta}( w \mid x ; \kappa, \mathcal{V}_{mle}, \mathcal{R}_{mle})\) appears to have a maximum value of 1 (i.e., the approximate MMI criterion places all the probability mass on the reference transcription \(w\)) when \(\theta\) is at a point at infinity.  This means that estimating model parameters using
\begin{equation*}
\hat{\theta} = \arg \max_{\theta \in \Theta} p_{\theta}( w \mid x ; \kappa, \mathcal{V}_{mle}, \mathcal{R}_{mle})
\end{equation*}
appears to be an ill-posed problem.  This also means that using \(p_{\theta}( w \mid x ; \kappa, \mathcal{V}_{mle}, \mathcal{R}_{mle})\) as a model selection criterion will produce estimates \(\hat{\theta}\) far away from \(\theta_{mle}\) where (\ref{mmiCritApprox}) is no longer a good approximation. 

Our explanation for the behavior of lattice-based MMI is in terms of the quality of the approximation (\ref{mmiCritApprox}) and the properties of \(p_{\theta}( w \mid x ; \kappa, \mathcal{V}_{mle}, \mathcal{R}_{mle})\) as a model selection criterion.  For the first several iterations of extended Baum-Welch, the estimated parameters, \(\theta_k\), stay close enough to \(\theta_{mle}\) so that the approximation (\ref{mmiCritApprox}) is still good.  As we have shown, parameter estimation using a model selection criterion close to \( p_{\theta}( w \mid x ; \kappa)\) is well behaved, and the first several iterations of model estimation using the approximate MMI criterion \( p_{\theta}( w \mid x ; \kappa, \mathcal{V}_{mle}, \mathcal{R}_{mle})\) is well behaved too: WERs on the training and test set steadily decline, while the approximate MMI criterion steadily increases.  However as the number of iterations of extended Baum-Welch increases beyond 20, we have shown that the approximation (\ref{mmiCritApprox}) degrades and the distributions \( p_{\theta}( \cdot \mid x ; \kappa)\) and \(p_{\theta_k}( \cdot \mid x ; \kappa, \mathcal{V}_{mle}, \mathcal{R}_{mle})\) become increasingly unrelated.  The WER on the training data performed by using the distribution \(p_{\theta_k}( \cdot \mid x ; \kappa, \mathcal{V}_{mle}, \mathcal{R}_{mle})\) (method B in Section~\ref{fixedLats}) steadily declines and the related approximate MMI criterion steadily increases, which shows that extended Baum-Welch is correctly optimizing the approximate MMI criterion.  However, the distribution that we really care about is not \(p_{\theta_k}( \cdot \mid x ; \kappa, \mathcal{V}_{mle}, \mathcal{R}_{mle})\), but rather what is now the very different distribution \(p_{\theta_k}( \cdot \mid x ; \kappa)\).  The WER on the training data using the distribution \( p_{\theta}( \cdot \mid x ; \kappa)\) (method A in Section~\ref{fixedLats}) begins to get steadily worse as do WERs on the test data.  This is not due to 'over fitting', but instead the natural consequence of using an ill-behaved approximation.

We also believe that this is the explanation for the behavior that was labeled `over fitting' in all the earlier versions of MMI.  Except in the case of tasks like isolated digits, it has never been possible to sum over all possible transcriptions or state sequences which the definition of the earlier MMI criterion, \(p_\theta(w \mid x)\), also requires.  Instead, all the early versions of MMI used approximations to this MMI criterion, say \(p_{\theta}( w \mid x ; \mathcal{V}_{mle}, \mathcal{R}_{mle})\), analogous to the approximation that we have studied in this paper.  It would be very surprising if the earlier approximations exhibited markedly different properties than those that we have shown \(p_{\theta}( w \mid x ; \kappa, \mathcal{V}_{mle}, \mathcal{R}_{mle})\) exhibits.

One approach to improving lattice-based MMI would be to come up with a better approximation to \(p_{\theta}( w \mid x ; \kappa)\). There are two directions that one could pursue.  The first would be to come up a better functional form for the approximation itself.  This could involve something as simple as adding a regularization term to prevent the model parameters from straying too far from the mle or a deeper analysis of what useful properties \(p_{\theta}( w \mid x ; \kappa)\) has that can be captured in practical form.  The second would involve a better, quantitative understanding of how close \(\theta\) needs to be to \(\theta_{mle}\) in order for the approximation (\ref{mmiCritApprox}) to be valid.  With that knowledge, instead of regenerating the lattices between each iteration of extended Baum-Welch, we could regenerate the lattices only as needed based on how far the model parameters have moved and our estimate of how much  (\ref{mmiCritApprox}) has deteriorated.  A related issue is that it would be extremely useful to have a quantitative measure of how `good' the lattices need to be for (\ref{mmiCritApprox}) to be valid.  Although \cite{povey2} contains some useful guidance in this regard, this remains part of the `black art' of lattice-based MMI.

Finally, we believe that this paper illustrates a fruitful area of research that is usually ignored in the speech recognition literature.  For a variety of reasons, we tend to treat our estimation algorithms as `black boxes'.  A great deal of attention is paid to the external responses of the algorithm to stimuli in the form of training and test data.  Especially important responses to these stimuli are computational efficiency and word error rate.  In fact, we tend to develop algorithms expressly to lower the word error rate, often by tinkering with or combining extant algorithms.  We also tend to prefer one algorithm over another based on these considerations, which leads to a form of word error rate and computational efficiency based natural selection in the evolution of algorithms.  While this state of affairs is perfectly reasonable, indeed it has led to remarkable progress in field of speech recognition over the last twenty years, what goes on inside the resulting rather intricate collection of black boxes has rarely if ever been examined.  For example, the models and their parameters typically live deep inside the black box, which means that asymptotic properties of the algorithm at the model parameter level are also hidden from view.  Another example is the overall effect of the many approximations that are used to make computations feasible.  When taken individually these effects are well understood, but their overall effect in combination is often overlooked or even ignored when theoretical analysis is undertaken.  Lattice-based MMI is a perfect example of one of these black boxes.  It is a complicated algorithm with many facets that has evolved over more than twenty years, the evolution driven by the goal of lowering the word error rate.  When considered in the light of word error rate reduction, lattice-based MMI is remarkably successful.  However, this paper shows that when we peer deeper inside this particular black box we see that the model parameters are behaving pathologically.  We were not only able to diagnose the problem -- a flaw in a key approximation -- but we were also able to correct it, which results in more stable model parameter behavior and a small improvement in the word error rate.  We intend to continue investigations of this sort on other important algorithms used in speech recognition.

\appendix

\section{What happens to the model parameters?} \label{modelParameters}

This Section presents further analysis of the results in Section~\ref{fixedLats}.  We will develop a simple framework that we will use to examine what happens to the model parameters after 100 iterations of extended Baum-Welch.  We shall demonstrate that extended Baum-Welch expands the space that the model means occupy and that this expansion appears to be consistent with steady decrease in the sequences \((\num(x, w ; \theta_k, \kappa, \mathcal{V}_{mle}, \mathcal{R}_{mle}))\) and \((\den(x, w ; \theta_k, \kappa, \mathcal{V}_{mle}, \mathcal{R}_{mle}))\) that we saw in Section~\ref{fixedLats}.  We also show that extended Baum-Welch appears to be shrinking the model variances, but not in a dramatic way.

We begin with a review of some useful properties of ellipsoids in \(\mathbb{R}^d\).  Let \(\Sigma\) be a \(d \times d\) positive definite matrix.  Then the inverse matrix \(\Sigma^{-1}\) defines a metric, \(d_{\Sigma^{-1}}(\cdot, \cdot)\),  on \(\mathbb{R}^d\) via
\begin{equation*}
d_{\Sigma^{-1}}(x, y) = \sqrt{(x-y)^t \Sigma^{-1} (x - y)}.
\end{equation*} 
This is called the \emph{Mahalanobis distance}.  The set of points in \(\mathbb{R}^d\) satisfying \(d_{\Sigma^{-1}}(x, 0) = 1\) is an ellipsoid
\begin{equation*}
Ell(\Sigma^{-1}) \equiv \{x \in \mathbb{R}^d : x^t \Sigma^{-1} x = 1\},
\end{equation*}
with volume given by
\begin{equation*}
\textrm{Volume}(Ell(\Sigma^{-1})) = \textrm{Volume}(S^{d-1}) \sqrt{\det \Sigma},
\end{equation*}
where \(S^{d-1} = \{x \in \mathbb{R}^d : \|x\| = 1\}\).
Let \(\{\lambda_i\}_{i=1}^d\) and \(\{u_i\}_{i=1}^d\) be the eigenvalues and corresponding choices for orthonormal eigenvectors for \(\Sigma\), where for convenience we have indexed the eigenvalues so that
\begin{equation*}
 \lambda_1 \leq \lambda_2 \leq \dots \leq \lambda_d.
\end{equation*}
Then the collection of vectors \(\{\sqrt{\lambda_i} u_i\}_{i=1}^d\) are the semi-principal axes of \(Ell(\Sigma^{-1})\).  The ratio \(c (\Sigma) \equiv \sqrt{\lambda_d / \lambda_1}\) gives a sense of how elongated the ellipsoid \(Ell(\Sigma^{-1})\) is: \(\sqrt{\lambda_d / \lambda_1} \geq 1 \) with equality \(\iff\) \(Ell(\Sigma^{-1}) = S^{d-1}\).\footnote{The quantity \(c^2\) is the \emph{condition number} of the matrix \(\Sigma\).  Also \(\sqrt{1 -  1 / c^2}\) is the \emph{eccentricity} of the two-dimensional ellipse formed by the intersection of \(Ell(\Sigma^{-1})\) with the plane through the origin spanned by \(u_1\) and \(u_d\).}

\begin{figure}
\centering
\includegraphics[width=.75\textwidth]{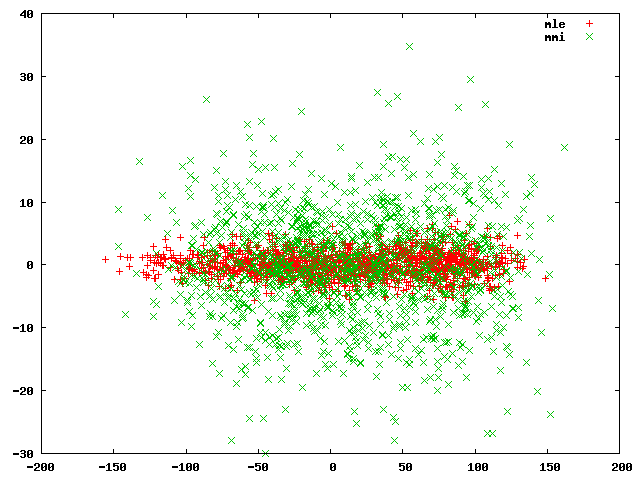}
\caption{Mean scatter along the top and bottom eigenvectors before and after 100 iterations of extended Baum-Welch.\label{meandist}}
\end{figure}

First we examine the scatter of the collection of the 1500 state means in \(\mathbb{R}^{39}\).  We treat these means as points in \(\mathbb{R}^{39}\) and compute the total mean vector (or centroid)
\begin{equation*}
\bar{\mu} = \frac{\sum_{j=1}^{1500} \mu_j}{1500},
\end{equation*}
and total variance or scatter matrix\footnote{Note that the scatter matrix for the means, \(T\), is the similar to the between class variation matrix.}
\begin{equation*}
T = \frac{\sum_{j=1}^{1500} (\mu_j - \bar{\mu}) (\mu_j - \bar{\mu})^t} {1500}.
\end{equation*}
Let \(T_{MLE}\) be computed using the mle means and \(T_{MMI}\) be computed using the means after 100 iterations of extended Baum-Welch.  The difference in the volume and elongation in the ellipsoids \(Ell(T_{MLE}^{-1})\) and \(Ell(T_{MMI}^{-1})\) is a crude measure of the effect 100 iterations of extended Baum-Welch has on the means.  The centroid does not move appreciably, but the volume that the means occupy increases by a factor of \(2.6 \times 10^5\).  Also, the measure of elongation starts out at \(c(T_{MLE}) = 950\) but after 100 iterations of extended Baum-Welch is decreases significantly to \(c(T_{MMI}) =  57\).  We would also like to inspect visually what is happening to the means.  Let's start with the mle.  We can project the 39-dimensional cloud of mle means onto the 2-dimensional plane spanned by the eigenvectors \(u_1(MLE)\) and \(u_{39}(MLE)\) for \(T_{MLE}\).  We can do the same thing with the means after 100 iterations of extended Baum-Welch, but this time projecting onto the different plane spanned by \(u_1(MMI)\) and \(u_{39}(MMI)\) for \(T_{MMI}\).  Figure \ref {meandist} compares these two projections.  Even though these projections are onto different 2-dimensional planes in \(\mathbb{R}^{39}\), we can still see that after 100 iterations of extended Baum-Welch:
\begin{itemize}
\item[1.] The spread in the cloud of means does not change appreciably in the direction of maximum variation (along the major axis).
\item[2.] The cloud of means becomes more spherical by becoming more spread out in the direction of the minimum variation (along the minor axis).
\end{itemize}
Note that that the cloud of mle means in Figure \ref {meandist} gives a sense of where the cloud of training data live.  It appears from Figure \ref {meandist} that 100 iterations of extended Baum-Welch has, in the direction of the minimum variation, pushed a significant fraction of the means well beyond the location of the training data.

We conclude that one effect of MMI is an expansion of the space that the model means occupy.  We also speculate that the direction of minimum variation in the model mean scatter is the least useful for discrimination, i.e. recognition, since it it also the direction of minimum variation in the between class scatter.  If this speculation is correct, then MMI is pushing apart the means the most in the direction that matters the least to recognition.  

\begin{figure}
\centering
\includegraphics[width=.75\textwidth]{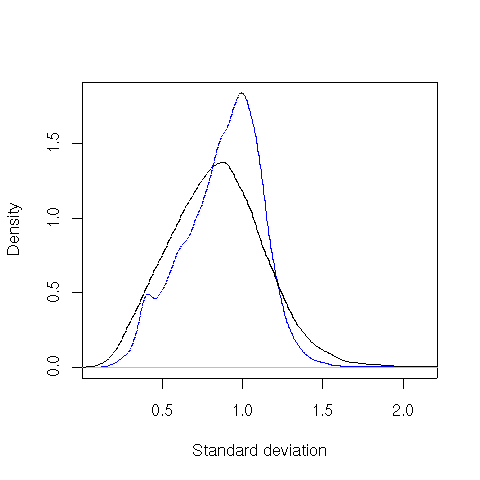}
\caption{Densities of the scaled standard deviations before (blue) and after (black) 100 iterations of extended Baum-Welch.\label{sigmadensity}}
\end{figure}

Next we examine what happens to the model variances after 100 iterations of extended Baum-Welch.  Recall we are using diagonal covariance matrices in our normal output distributions.

To start with we shall examine histograms of the model standard deviations before and after extended Baum-Welch.  Instead of creating 39 separate histograms, one for each of the features, would like to create just one histogram that combines the model standard deviations in all of the 39 feature dimensions.  To do this we rescale each feature so that its total variation on the training data is 1, and rescale the standard deviations accordingly.  After this rescaling, the deviations is different feature dimensions are roughly commensurable.  Figure \ref{sigmadensity} displays densities fit to the resulting histograms.  It appears that the standard deviations have shrunk somewhat after 100 iterations of extended Baum-Welch, although the effect is not dramatic.  It is worth noting that none of the mle model variances were floored, while only 10 out of the approximately 59,000 variances are floored after 100 iterations of extended Baum-Welch.

\begin{figure}
\centering
\includegraphics[width=1\textwidth]{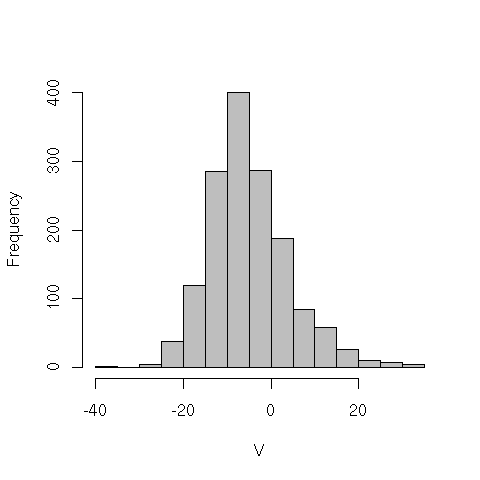}
\caption{Histogram of \(\{V_j\}_{j=1}^{1500}\).\label{varhist}}
\end{figure}

For each state \(j\) let \(\sigma_{mle, j}\) and \(\sigma_{mmi, j}\) denote the vectors of standard deviations for state \(j\) before and after 100 iterations of extended Baum-Welch, and define
\begin{equation*}
V_j \equiv \log \frac{\prod_{i=1}^{39}\sigma_{mmi, j, i} }{\prod_{i=1}^{39}\sigma_{mle, j, i} }.
\end{equation*}
For each \(j\) the quantity \(\prod_{i=1}^{39}\sigma_{mle, j, i}\) gives the volume of the 39-dimensional rectangular parallelepiped with edges specified by the elements of the vector \(\sigma_{mle, j}\); this is also proportional to the volume of the ellipsoid \(Ell(\diag(\sigma^2_{mle, j}))\).  So the quantity \(V_j\) measures how this volume changes after 100 iterations of extended Baum-Welch.  Figure \ref {varhist} displays a histogram of the collection \(\{V_j\}_{j=1}^{1500}\), three quarters of which are negative.

Recall that, modulo constants, the log-likelihood of a frame \(y \in \mathbb{R}^{39}\) with respect to \(N(\mu, \sigma^2)\) is
\begin{equation*}
S(y) = - \frac{1}{2} \sum_{i=1}^{39} \frac{(y_i - \mu_i)^2}{\sigma^2_i} - \log \prod_{i=1}^{39}\sigma_i,
\end{equation*}
and that the numerator and denominator log-likelihoods are dominated by state specific versions of \(S\).  Moreover, Figure \ref {numdenebw} in Section shows that after 100 iterations of extended Baum-Welch
\begin{equation} \label{morenegative}
\den(x, w ; \theta_{100}, \kappa, \mathcal{V}_{mle}, \mathcal{R}_{mle}) < \den(x, w ; \theta_{mle}, \kappa, \mathcal{V}_{mle}, \mathcal{R}_{mle}).
\end{equation}
The expansion in the model means that we have observed is consistent with (\ref{morenegative}), since it appears that in at least some dimensions the means are moving away from the data which would make the first sum in terms like \(S\) more negative.  However, the our crude analysis of the behavior of the variances is mixed relative to (\ref{morenegative}).  On the one hand the standard deviations appear to be shrinking after 100 iterations of extended Baum-Welch, which would again make the first sum in terms like \(S\) more negative, but on the other hand we saw that the volume terms also appear to be shrinking which would make the second sum in terms like \(S\) less negative. 

\section{Experiments with \(E\)} \label{E_experiments}

In this Section we explore what effect the extended Baum-Welch parameter \(E\) has on the results in Section~\ref{fixedLats}. The parameter \(E\) controls two properties of extended Baum-Welch. On the one hand, if \(E\) is large enough, then each iteration of extended Baum-Welch will result in an increase in the MMI criterion (\cite{kanevsky}, \cite{axelrod}).  On the other hand, \(E\) may be thought of as a smoothing parameter that controls how related the parameters \(\theta_{k+1}\) are to the previous parameters \(\theta_{k}\).  A small value for \(E\) will produce a larger difference \(\| \theta_{k+1} - \theta_k \|\) than a large value for \(E\) will produce.  This has led to the belief that choosing \(E\) large enough not only ensures the convergence of the sequence of approximate MMI criteria \((p_{\theta_k}( w \mid x ; \kappa, \mathcal{V}_{mle}, \mathcal{R}_{mle}))\) but also ensures the convergence of the underlying sequence of parameters \((\theta_k)\).

\begin{figure}
\centering
\includegraphics[width=1.0\textwidth]{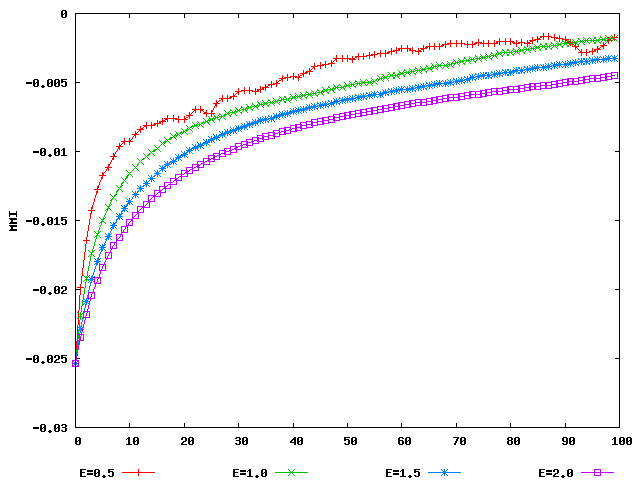}
\caption{The MMI criterion on the training data during 100 iterations of extended Baum-Welch using different values for \(E\). The x-axis gives the extended Baum-Welch iteration, with x=0 being the mle.\label{Esweep_mmi}}
\end{figure}

We repeat the experiment that used fixed phone-marked word lattices generated by the mle and \(E = 1.0\) described in Section~\ref{fixedLats} but this time with three other choices for \(E\).  Figure \ref{Esweep_mmi} shows how the MMI criterion varies as extended Baum-Welch proceeds when \(E\) is set to 0.5, 1.0, 1.5, and 2.0.  Note that \(E = 0.5\) is too small to ensure that the approximate MMI criterion increases at each iteration of extended Baum-Welch.  Note also that as the value of \(E\) increases the corresponding trajectories in the approximate MMI criterion become smoother.  Finally, note that the trajectory in the approximate MMI criterion when \(E=2\) looks remarkably similar to the corresponding trajectory when \(E=1.0\), albeit smoother and shifted to the right.  

\begin{figure}
\centering
\includegraphics[width=1.0\textwidth]{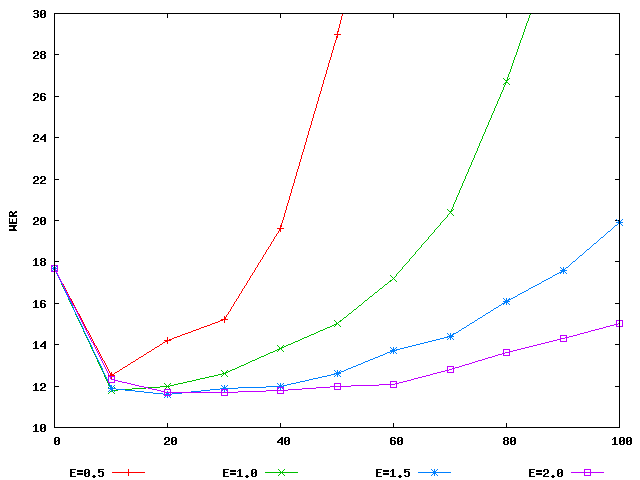}
\caption{The WER measured on the test set during 100 iterations of extended Baum-Welch using different values for \(E\). The x-axis gives the extended Baum-Welch iteration, with x=0 being the mle.\label{Esweep_wer}}
\end{figure}

Figure \ref{Esweep_wer} and Table \ref{Esweep_wer_table} show how the WER on the independent test varies as extended Baum-Welch proceeds when \(E\) is set to 0.5, 1.0, 1.5, and 2.0.  Larger values of \(E\) appear to be just delaying the inevitable rise in the sequence of WERs.  When \(E\) is set to 0.5, 1.0, or 1.5, the WER at iteration 100 is larger than that of the MLE.  We also ran 100 more iterations of extended Baum-Welch when \(E=2.0\).  In this case the WER on the independent test set is \(42.8\%\) at iteration 200, which is nearly the same as the WER when \(E=1.0\) at iteration 100, namely \(42.2\%\).  Roughly speaking, as the reader can verify from Figure \ref{Esweep_wer} or Table \ref{Esweep_wer_table}, the WER at iteration \(k\) with \(E=e_1\) will be the same at iteration \(\frac{e_2}{e_1} k\) with \(E=e_2\).   

\begin{table}
\centering
\begin{tabular}{|r|c|c|c|c|}
\hline
& \multicolumn{4}{|c|}{E}\\ \cline{2-5}
iteration & 0.5 & 1.0 & 1.5 & 2.0\\
\hline
\hline
mle &   17.7 &   17.7 &   17.7 &   17.7\\
10 &    12.5 &   11.8 &   11.9 &   12.3\\
20 &    14.2 &   12.0 &   11.6 &   11.7\\
30 &    15.2 &   12.6 &   11.9 &   11.7\\
40 &    19.6 &   13.8 &   12.0 &   11.8\\
50 &    29.0 &   15.0 &   12.6 &   12.0\\
60 &    38.6 &   17.2 &   13.7 &   12.1\\
70 &    45.3 &   20.4 &   14.4 &   12.8\\
80 &    49.7 &   26.7 &   16.1 &   13.6\\
90 &    54.6 &   34.4 &   17.6 &   14.3\\
100 &   52.2 &   42.2 &   19.9 &   15.0\\
\hline
\hline
\end{tabular}
\caption{WER on the test set during 100 iterations of extended Baum-Welch using different values for \(E\).\label {Esweep_wer_table}}
\end{table}

While choosing \(E \geq 1\) does ensure that each iteration of extended Baum-Welch results in an increase in the approximate MMI criterion, values of \(E\) larger than 1 only appear to delay the pathological behavior that we observed in Section~\ref{fixedLats}.

\section{MPE}\label{mpe}

Minimum phone error (MPE) training (\cite{povey2}) uses yet another model selection criterion coupled with an iterative estimation algorithm that is essentially the same as extended Baum-Welch.  Given \(\textrm{w} \in \mathcal{W}\) then we define \(A(\textrm{w}, w)\) to be the phone accuracy of the transcription \(\textrm{w}\) measured relative to the true transcription \(w\) and a dictionary that provides phone level pronunciations for each transcription in \(\mathcal{W}\).  Then the MPE criterion, \(\mathcal{F}_{MPE}\), is defined at each \(\theta\) to be
\begin{equation*}
\mathcal{F}_{MPE}(\theta) = \sum_{\textrm{w} \in \mathcal{W}} p_\theta(\textrm{w} \mid x; \kappa) A(\textrm{w}, w).
\end{equation*}

In this Section we report preliminary results on an experiment analogous to the experiment in Section~\ref{fixedLats}: we follow the standard procedure in the literature by generating phone-marked word lattices once and all using the mle, but this time we use an approximation to the MPE criterion rather than an approximation to the MMI criterion for model selection during 100 iterations of extended Baum-Welch.  Since we will use fixed phone-marked word lattices for each pass of extended Baum-Welch, this means that the corresponding approximate MPE criterion that we use for model selection is given by:
\begin{equation*}
\mathcal{F}_{MPE}(\theta; \kappa, \mathcal{V}_{mle}, \mathcal{R}_{mle}) = \sum_{\textrm{v} \in \mathcal{V}_{mle}} p_\theta(\textrm{v} \mid x; \kappa, \mathcal{V}_{mle}, \mathcal{R}_{mle}) A(\textrm{v}, w).
\end{equation*}

\begin{figure}
\centering
\includegraphics[width=1.0\textwidth]{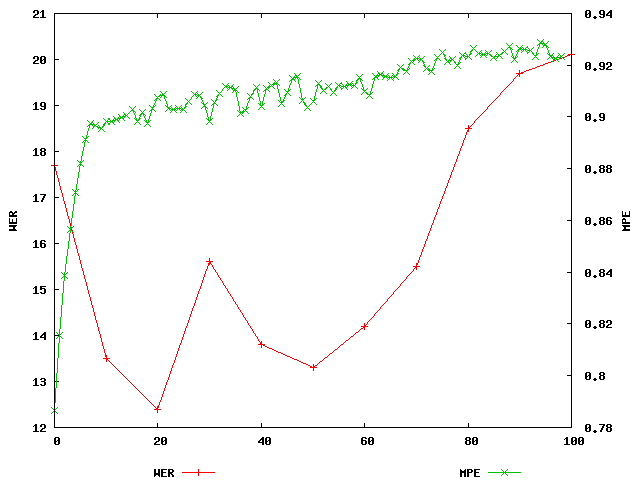}
\caption{The WER on an independent test set and the MPE criterion on the training data during 100 iterations of extended Baum-Welch.  The x-axis gives the extended Baum-Welch iteration, with x=0 being initial models.\label {mpe_graph}} 
\end{figure}

Figure \ref{mpe_graph} displays the WER on the independent test set and the approximate MPE criterion as extended Baum-Welch proceeds.\footnote{We use HTK to perform this experiment with standard settings for MPE, namely \(E = 2.0\) and \(\tau = 50\).} Since Figure \ref{mpe_graph} only displays the WER every tenth iteration, we note that the WER reaches a minimum WER of \(11.7\%\) on iteration 6, which the reader will recall is similar to the performance that we observed in Section~\ref{fixedLats} using MMI. 

One might expect MPE to be less prone to the problems that we observed with MMI in Section~\ref{fixedLats}, since the MPE criterion is in a sense self-regulating: if we push the model parameters, \(\theta\) too far away from the mle, then presumably this will induce phone level errors which will result in a lower value for \(\mathcal{F}_{MPE}(\theta; \kappa, \mathcal{V}_{mle}, \mathcal{R}_{mle})\).  Unfortunately, this self-regulation is not sufficient to prevent the pathological behavior exhibited in Figure \ref{mpe_graph}.  The overall trend in the approximate MPE criterion is upward, but, after iteration 50 so is the overall trend in test set WER.

\end{document}